\documentclass[sigconf,10pt]{acmart}

\renewcommand\footnotetextcopyrightpermission[1]{} 
\usepackage{listings}
\usepackage{graphicx}
\usepackage{multirow}
\usepackage{balance}

\AtBeginDocument{%
  \providecommand\BibTeX{{%
    \normalfont B\kern-0.5em{\scshape i\kern-0.25em b}\kern-0.8em\TeX}}}

\begin{document}

\title{A Scalable Space-efficient In-database Interpretability Framework for Embedding-based Semantic SQL Queries}

\author{Prabhakar Kudva}
\email{kudva@us.ibm.com}
\affiliation{%
  \institution{IBM T.J. Watson Research Center}
  \city{Yorktown Heights}
  \state{NY}
  \country{USA}}

\author{Rajesh Bordawekar}
\email{bordaw@us.ibm.com}
\affiliation{%
  \institution{IBM T.J. Watson Research Center}
  \city{Yorktown Heights}
  \state{NY}
  \country{USA}}

\author{Apoorva Nitsure}
\email{apoorva.nitsure@ibm.com}
\affiliation{%
  \institution{IBM T.J. Watson Research Center}
  \city{Yorktown Heights}
  \state{NY}
  \country{USA}}

\begin{abstract}

AI-Powered database (AI-DB) is a novel relational database system
that uses a self-supervised neural network, database embedding, to enable semantic
SQL queries on relational tables. In this paper, we describe an architecture and implementation of
in-database interpretability infrastructure designed to provide
simple, transparent, and relatable insights into ranked results of semantic SQL queries supported by
AI-DB. We introduce a new co-occurrence based interpretability 
approach to capture relationships between relational entities and
describe a space-efficient probabilistic Sketch implementation to
store and process co-occurrence counts. Our approach provides both query-agnostic
(global) and query-specific (local) interpretabilities. Experimental evaluation
demonstrate that our in-database probabilistic approach provides the same
interpretability quality as the precise space-inefficient approach, while 
providing scalable and space efficient runtime behavior (upto 8X space savings), without any user intervention.

\end{abstract}

\maketitle
\pagestyle{plain}

\section{Introduction}
\label{sec:intro}
AI-Powered Database (AI-DB) is a novel approach for enabling
semantic capabilities in relational databases by using a
\emph{built-in} self-supervised neural
network model~\cite{bordawekar:corr-abs-1603-07185, Bordawekar:deem17, spark-cogdb}. AI-DB is built on 
an observation that there is a large amount of \emph{untapped hidden information} in a database relation, mainly in the form of inter-/intra-column, 
and inter-row relationships. Further, current SQL queries, various text
extensions, and user-defined functions (UDFs) \emph{are not
sufficient} to retrieve  
this information. AI-DB employs a self-supervised neural network
called, database embedding (db2Vec), to capture this latent
information from relational
tables~\cite{Bordawekar:deem17}.  For a relational
table, db2Vec builds a semantic model that captures inter- 
and intra-column relationships between entities of different types
(db2Vec is governed by the relational data
model~\cite{DBLP:journals/sigmod/Date82}, not by a language model). After training, AI-DB seamlessly integrates
the trained model into the existing SQL query processing infrastructure and
uses it to enable a new class of SQL-based semantic queries called
Cognitive Intelligence (CI) queries. The CI queries use the trained
embedding vectors to extend the traditional \emph{value-based} SQL query
capabilities of the relational data model to enable \emph{semantic}
operations on relational entities. The output of a SQL CI query is a ranked 
list of rows that is ordered by a relevance similarity score. Both Spark~\cite{spark-cogdb} and native
Python implementations of AI-DB support a variety of CI queries, including
similarity, dissimilarity, and inductive reasoning queries such as analogies 
or semantic clustering, and semantic grouping~\cite{neves:aidb19}. 
 The recently released IBM$\textsuperscript{\textregistered}$ Db2$\textsuperscript{\textregistered}$ 
13 for z/OS$\textsuperscript{\textregistered}$ SQL Data 
 Insights~\cite{db2-sdi} is an implementation of AI-DB in
 Db2$\textsuperscript{\textregistered}$ 13 for
 z/OS$\textsuperscript{\textregistered}$ on the IBM
 zSystems$\textsuperscript{\textregistered}$ platform~\cite{ibm-zADE}.  

Given that the semantic operations are based on the
meanings inferred during the database embedding training process, it is
imperative that the end user understands the dominant factors that 
impact the result rankings. Describing result rankings in terms of just database
embeddings can be opaque and inscrutable to a customer. The main goal of this work
is to provide \emph{simple, transparent}, and \emph{relatable} result insights that are accessible to
traditional database users, who may not be familiar with the intricacies of machine learning (ML) approaches.
The key to
understanding how the database embedding works is to get details on the
co-occurrence statistics of the input data and expose their influence of the 
result rankings. Since any relational
table has a large number of entities of different types, it is not a
trivial task to explore such a large amount of statistical information. 

This paper describes the design and implementation of a scalable,
space-optimized intepretability infrastructure in AI-DB
(Figure~\ref{fig:poverview}). Key novelties from our work include:

\begin{itemize}

\item A new approach that enables both query-agnostic (global) and
  query-specific (local) interpretability for the self-supervised
  database embedding model, db2Vec.

\item Support for interpreting ranked results from a semantic similarity
  function by identifying dominant entities from the input relational table
  
\item Implementing hooks in the db2Vec code to extract model-intrinsic
  token occurrence statistics.

\item Introduction of two numeric metrics, Influence and Discriminator scores, to 
enable local model interpretability as well as capturing input data characteristics

\item Using a probabilistic in-memory data structure, Count-Min sketch, to store
  the co-occurrence counts, followed by using a sparse matrix storage format
  to generate its serialized representation, and then populating an auxiliary relational
  table to be used at runtime for global function interpretability.

\item Integration of the interpretability capability into the
  end-to-end AI-DB pipeline.

\end{itemize}

Although the co-occurrence based interpretabilty has been currently designed for
AI-DB, it has broader applicability for understanding results from
querying any context-based vector embedding models such as Large Language Models (LLMs) designed 
for different domains such as text, chemistry, or coding~\cite{nvidia-llm, hugface, openai-text}.

The rest of the paper is organized as follows. Section~\ref{sec:related} 
describes the notion of interpretability, and discusses recent related
work. Section~\ref{sec:aidb} first presents an overview of AI-Powered Database
and semantic SQL CI queries and then discusses how interpretability is
used for providing insights into AI-DB's SQL CI
queries. Section~\ref{sec:impl} introduces two intepretability approaches and details implementation of
a space-efficient data structure based on the Count-Min
sketch. Section~\ref{sec:eval} presents experimental evaluation of the
sketch data structure and its usage in interpreting AI-DB semantic
operations. Finally we conclude in Section~\ref{sec:concl}.

\section{Related Work}
\label{sec:related}


Understanding the entire lifecycle of machine learning or deep learning usage, from 
preparing the training data, setting up the training instance, and then using the trained model
to get results, is a very active area of investigation towards creating \emph{responsible} AI. 
The main goal of these activities is
provide users with different ways to understand, assimilate, and develop trust in the results generated by
the trained AI model. 

A majority of current explorations towards understanding the behavior of
an AI model are targeted towards \emph{supervised} models
designed to predict a \emph{target} class for an unseen value (classification). 
In literature, such work on understanding AI models is often referred to as \emph{interpretability} or
\emph{explainability}~\cite{molnar2022,arrieta:xai,
  DBLP:journals/corr/abs-2010-00711,DBLP:journals/corr/abs-1806-00069, ricards:zoo}.
Typically, \emph{explainability} refers to techniques that aim to provide insights into results
from a \emph{black box} model via creating a second \emph{post hoc}
model~\cite{rudin:intp}, whereas \emph{interpretability} refers to \emph{white box} techniques that 
use \emph{model-intrinsic} information to provide result
insights. These techniques can be further characterized by the scope of their
inquiries: \emph{global} approaches provide
\emph{query-agnostic} insights on the overall structure, parameters, and behavior
of a model, whereas \emph{local} approaches help understand how a
model makes decisions for an individual prediction
query~\cite{du:cacm-intp}.

Traditional machine learning (ML) models such as
generalized linear models (GLMs), decision trees, or tree-based
ensemble models use feature engineering to map raw data
into features. In contrast, deep learning (DL) models, such as the
embedding models used in this work, learn \emph{representations} from
raw data. In the case of ML models, one can use a measure called \emph{feature importance}
to capture the statistical contribution of each feature into the 
overall decision making process to enable global interpretability. For simple ML approaches such as GLM or
decision trees, it is often very easy to capture feature
importance. However, for tree-based ensemble models (e.g., random forests or gradient boosted machines
such as XGBoost or LightGBM), different techniques, e.g., feature
coverage, are needed to capture the contributions of each individual
feature~\cite{du:cacm-intp}. Other examples of global interpretability include exploiting
algorithmic internal information such as distribution plots, dependence plots which  
showcase the marginals and their interactions, permutation feature
importance which perturb values and measure error to rank
features. 

Examples of query-specific \emph{local} explanation
tools include LIME (Local Interpretable Model-Agnostic
Explanations)~\cite{ribeiro:lime} and SHAP (\underline{SH}apley
  \underline{A}dditive
  ex\underline{P}anations)~\cite{lundberg:shap}. LIME is a
  model-agnostic technique that, for any individual prediction query,
  approximates behavior of a black-box ML or DL model by using an
  interpretable surrogate model (e.g., a decision tree or a linear
  classifier with strong regularization). The outputs (predictions) of
  the surrogate model using a permuted sample input data are then
  computed. These predictions along with the information on the data
  permutations, provide insights on the contributions of each feature
  on the prediction of a data sample. SHAP is a game theory-based model agnostic local
interpretability method, which provides how every feature contributes to the
prediction, assuming each feature or group of features is a player in
game theory context. The SHAP algorithm 
  returns a Shapley value that expresses model prediction as a linear
  combination of binary variables that describe if a particular
  feature is present in the model or not. Both LIME and SHAP
  approaches work for multiple types of supervised ML and DL models.

Unlike supervised learning, understanding results from unsupervised or
self-supervised DL models (e.g., vector embedding models) has not
received enough attention. Traditional approaches such as LIME or SHAP
are not able to explain vector embedding models due to the complex
internal representations learned from the raw data~\cite{ho:dl-complexity}.
Recently there have been a few post hoc \emph{local} explainability proposals~\cite{madsen:survey} to understand the embedding model:
approaches such as SPINE~\cite{SPINE} and others~\cite{zhang:w2v,zhang:survey} rely
either on supplementing the embedding process by another network
architecture or create new embeddings using the existing embeddings as
input, while ~\cite{lex2vec} tries to explain the word embedding model
by exploiting external lexical resources. Allen et al.~\cite{allen:analogy}
explains analogy queries on a word embedding model using Pointwise
Mutual Information~\cite{church-hanks-1990-word}. Structural probing
covers a class of methods that use an external classifier (e.g., a
logistic regression) to learn a mapping from an internal
representation to a linguistic representation, e.g.,
BERTology~\cite{belinkov:survey, bertology, madsen:survey}. Unlike the
local explanation approaches, \emph{global post hoc} explanation approaches aim
to describe the entire model in form of relationships between the
words in the vocabulary. Usually such methods employ transformations
over the embedding matrix such as projections to two or three
dimensions, e.g., PCA~\cite{pca} or t-SNE~\cite{JMLR:v9:vandermaaten08a}, or
rotation~\cite{park-etal-2017-rotated}. Recently, the topic of whether the attention mechanism~\cite{bahdanau:arxiv}
 used in the transformer models~\cite{vaswani2017attention} is suited for explaination 
~\cite{DBLP:journals/corr/abs-1902-10186,DBLP:journals/corr/abs-1908-04626}
has been debated~\cite{bibal-etal-2022-attention} and has shown promise for some domains of application such as translation.

Overall, for self-supervised
DL models, currently most approaches are post hoc, focusing on
different aspects of the trained model. Currently, there is no single
post hoc approach or an interpretability approach that uses
model-intrisic information for providing detailed insights into the
behavior of various self-supervised DL models. None of these techniques apply for the AI-DB scenario as the
underlying model (db2Vec) is a specialized
self-supervised DL model, built over a relational table
with text and numeric entities (i.e., the training is governed by the
relational data model, not a language model), and the result is a list
of ranked rows, not a value to be predicted.

\section{AI-Powered Database Overview}
\label{sec:aidb}

\begin{figure}[htbp]
  \centering
  \includegraphics[width=\linewidth]{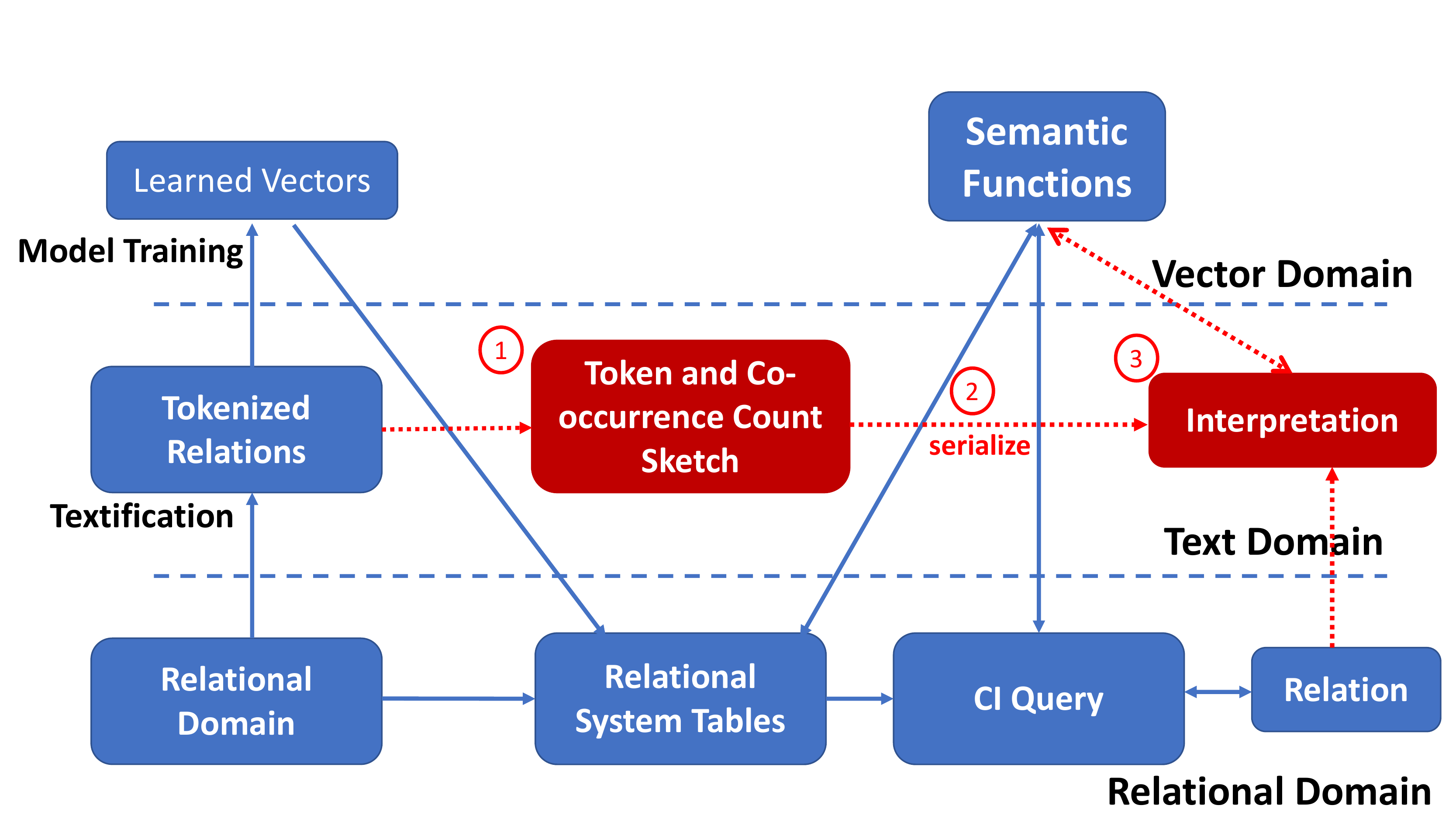}
  \caption{AI-Powered Database (AI-DB) execution stages. Operations are executed in three domains: relational, text, and vector.}
  \label{fig:poverview}
\end{figure}

Figure~\ref{fig:poverview} presents key stages in the execution flow of an AI-Powered Database (Figure~\ref{fig:poverview}). 
The first \emph{preprocessing phase} takes a relational table with different
SQL types as input and returns an unstructured but meaningful text corpus consisting of a set of English-like sentences. 
This transformation, termed \emph{textification}, allows us to generate a uniform semantic representation of  different SQL types. 
The textification phase processes each relational row separately and for each row, converts data of different 
SQL data types to equivalent text representation (\emph{tokens}). For example, numeric values are first clustered~\cite{bordawekar:corr-abs-1603-07185},
and then each numeric value is replaced by a string token that represents the corresponding numeric cluster (e.g., a value \texttt{121} 
is represented by a token \texttt{c3}, if \texttt{121} is mapped to the cluster 3.)

Once a relational table is converted into a textual training dataset,
it is processed by a self-supervised neural network, \emph{database embedding (db2Vec)}, to  
generate semantic vectors for every unique entity in the training
dataset (\emph{training phase})~\cite{bordawekar:corr-abs-1603-07185,Bordawekar:deem17}. The db2Vec differs from other NLP vector embedding models  
used for language modeling, such as Word2Vec~\cite{mikolov:nips13} or
GloVe~\cite{pennington:glove14} in several ways. Key differences
include: 

\begin{itemize}

\item A sentence generated from a relational row is 
not in any natural language such as English. In the db2Vec implementation, every token in the training set has
the same influence on the nearby tokens; i.e. the generated sentence is viewed as
\emph{an unordered bag of words}, rather than \emph{an ordered sequence}.

\item Unlike the untyped text document being used in NLP, the tokens in the textified training
document are \emph{typed} as per the associated column name in the relational table.

\item db2Vec builds a common vector-embedding model over a multi-modal relational table that
can contain entities of different types such as numeric values. The trained vectors are \emph{untyped}, thus
enabling semantic operations on relational values of different types.

\item For relational data, db2Vec provides special consideration to
  primary keys. Traditional word embedding approaches discard less frequent words
  from computations. In our implementation, by default, every token, irrespective
  of its frequency, is assigned a vector. For an unique primary key, its vector
  represents the meaning of the entire row.

\item The db2Vec model provides special treatment for the
  entities corresponding to the SQL \texttt{NULL} (or equivalent)
  values. The \texttt{NULL} values are processed such that they do not
  contribute to the meanings of neighboring non-null entities; thus
  eliminating false similarities. 

\end{itemize}

 For each 
database value in a relational table, the model generates a \emph{meaning} vector that encodes contextual
semantic relationships generated by collective contributions of other 
tokens within and across rows (each table row is viewed as a
sentence). The db2Vec model generates a variety of
semantic vectors: (1) each unique primary key is associated with a
semantic vector that captures behavior of the entire relational row
associated with that key, (2) for all other entities, their semantic
vectors capture collective contributions of their neighborhood
entities across their occurrences, and (3) table schema types (column
names) get their meaning vectors that capture table-wide
relationships. Once the model is trained, the \emph{model storage phase} stores 
the model into a relational systems table using a token as the key, and the corresponding
meaning vector as the value.  

 The final \emph{query execution phase} is where the user issues SQL
 statements to extract information from one or more databases using
 the trained db2Vec models. Such queries, termed
 \emph{Cognitive Intelligent (CI)} queries, can support both the traditional value-based 
as well as the new semantic contextual computations in the same
query~\cite{Bordawekar:deem17, neves:aidb19}. Each CI query uses special semantic functions to measure semantic similarity
between tokens associated with the
input relational parameters. The core computational operation of a semantic function is to calculate similarity
between a pair of tokens by computing the cosine distance
between the corresponding meaning vectors. For two vectors $v_1$ and
$v_2$, the cosine distance is computed as $cos(v_1 , v_2 ) =
cos(v_1,v_2)=\frac{v_1\cdot v_2}{\lVert {v_1} \rVert \lVert 
  {v_2} \rVert}$. The cosine distance value varies from 1.0 (very
similar) to -1.0 (very dissimilar). 

\begin{figure}[htbp]
\hrule
{\small \tt{
\begin{flushleft}
SELECT C.CUSTOMER\_NAME, AI\_SIMILARITY(C.CUSTOMER\_NAME,'CUST0')\\
AS similarityValue\\
FROM Customer\_View C\\
WHERE AI\_SIMILARITY(C.CUSTOMER\_NAME, ‘CUST0’) > 0.5\\
ORDER BY similarityValue DESC\\
\end{flushleft}
}
\hrule
}
\caption{Example of a SQL CI similarity query: find customers with similar overall behavior as other customers}
\label{fig:example}
\end{figure}

Figure~\ref{fig:example} presents a simple
CI SQL query that given a customer identifier, \texttt{CUST0}, identifies other
customers with the similar overall behavior (each row in the view \texttt{Customer\_View} 
represents a customer transaction). The semantic function, \texttt{AI\_SIMILARITY()}, takes
relational variables as input, and returns a similarity score computed via invoking cosine similarity over 
corresponding meaning vectors. The SQL CI query then returns those rows whose semantic similarity is
higher than a specified bound (0.5), ranked in a decreasing order of the similarity value. Customers 
whose similarity scores are closer to 1.0, are viewed to be semantically similar to the
input customer, \texttt{CUST0}. Irrespective of the SQL data types of input parameters, the similarity computations
are executed using untyped meaning vectors. This enables CI queries to invoke semantic functions on parameters of different 
SQL data types. Thus, unlike the supervised training model
that works only for the target entity it was trained for, a single database embedding model can be used 
for answering semantic queries over a value of any SQL data type from the associated relational table.

\subsection{Interpretability in AI-DB}
\label{sec:interpret}

Unlike the vector embedding models used in NLP, db2Vec trains an
input training document using the relational data model~\cite{DBLP:journals/sigmod/Date82}. db2Vec 
operates on a training document which is an untyped textified version of a relational table, and views it 
as a set of \emph{neighborhood contexts}, each defined by the
corresponding row in the input relational source. Each context is
viewed as an unordered bag of words, where each token is \emph{equally} related to other
tokens in the context (in contrast, NLP vector embedding techniques view a
training document as a \emph{sequence} of words whose relationship is
inversely proportional to the distance between the words). 

\begin{figure}[htbp]
  \centering
  \includegraphics[width=\linewidth]{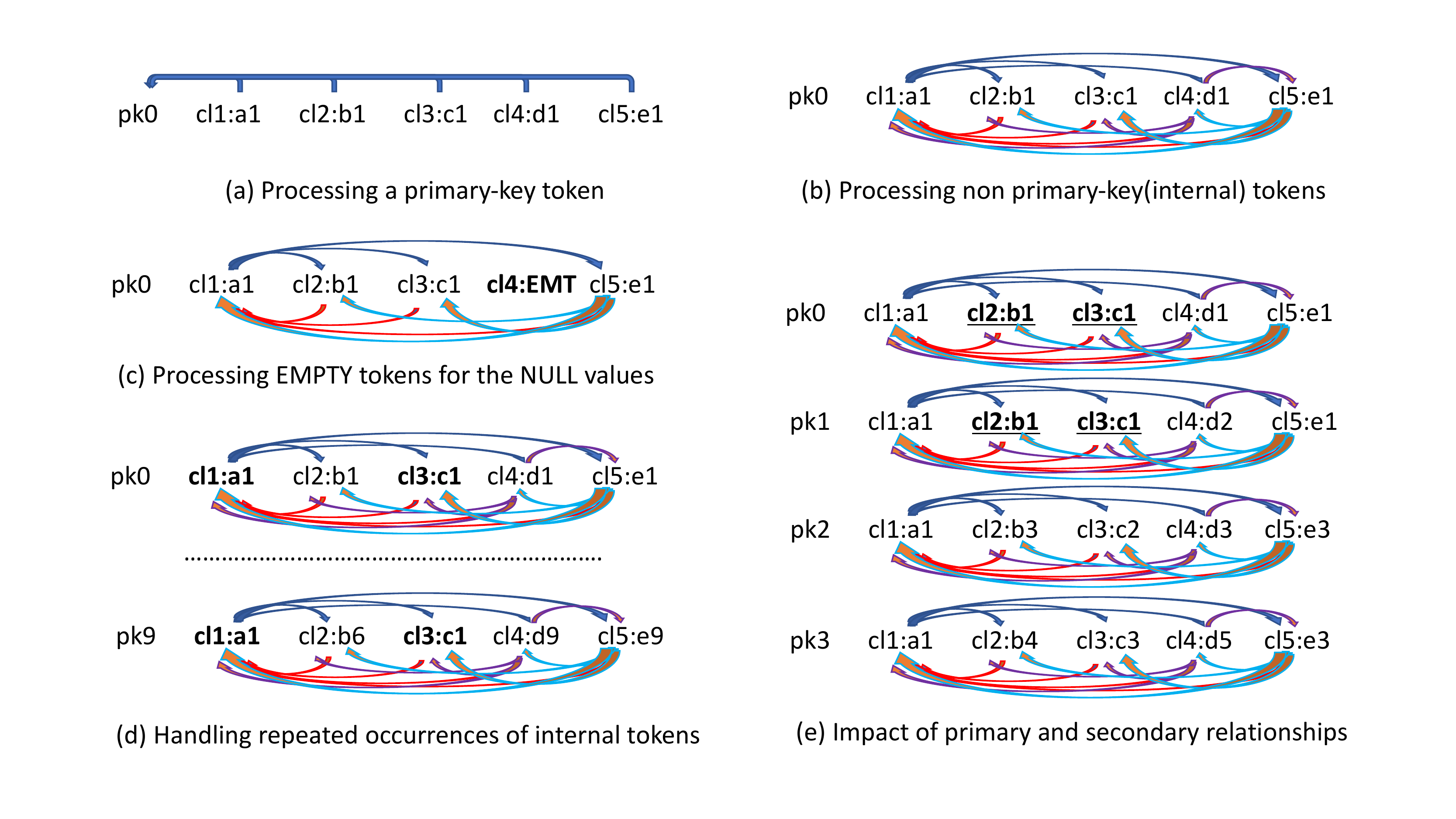}
  \caption{Capturing relationships in a database embedding model for primary (a), non-primary key (b), empty (c), repeated occurrences of tokens (d), and primary and secondary relationships between tokens (e)}
  \label{fig:db2vec}
\end{figure}

Each textified token is represented as a tuple, \texttt{type:value},
where \texttt{type} denotes the corresponding column name, and \texttt{value} represents
the textified representation of the associated relational entity.  From a
training perspective, db2Vec separates input tokens in three groups:
(1) primary-key tokens, (2) empty (\texttt{EMT}) tokens corresponding
to the SQL \texttt{NULL} values, (3) all other tokens
(Figure~\ref{fig:db2vec}). 
For those tokens representing unique
primary key values, their inferred meaning is generated using
contributions from \emph{internal} tokens only within its neighborhood (i.e.,
corresponding row in the relational source). The primary key token
does not contribute to the meaning of any other tokens in the training
document (Figure~\ref{fig:db2vec}(a)). All empty tokens corresponding
to SQL \texttt{NULL} values (e.g., \texttt{cl4:EMT}, Figure~\ref{fig:db2vec}(b)) are neglected during
training (i.e., no vectors are generated). For the remaining internal
tokens, within a neighborhood, each internal token contributes to the
meaning of other internal tokens and vice versa
(Figure~\ref{fig:db2vec}(c)). An internal token can appear multiple
times in a training document (e.g., \texttt{cl1:a1}, represents a relational value occurring
in multiple row); its final inferred meaning is the collective
contributions of all other internal tokens that co-occur in
multiple contexts (Figure~\ref{fig:db2vec}(d)). 

Using aforementioned rules,
db2Vec traverses the training document multiple times, iteratively
building the semantic model. As the training progresses, each token
becomes related to every other token in the training document, even if
the tokens do not co-occur. This can impact the relationships between
co-occurring tokens; for example, in Figure~\ref{fig:db2vec}(e),
although the token \texttt{cl1:a1} has the same co-occurrence count of 2
with tokens \texttt{cl5:e1} and \texttt{cl5:e3},  \texttt{cl1:a1} is
more closely related to \texttt{cl5:e1}, as they share the two
neighbors,  \texttt{cl2:b1}, and \texttt{cl3:c1}, with the same
co-occurrence count as \texttt{cl1:a1} and \texttt{cl5:e1}. Thus, the
\emph{primary} relationship between \texttt{cl1:a1} and \texttt{cl5:e1} gets
strengthened by the \emph{secondary} relationships of the shared neighbors.
At the end of training, db2Vec generates a semantic model containing real-valued
vectors of dimension, $d=320$, for every unique token in the input training document.

These vectors
are used by the semantic functions \newline (e.g., \texttt{AI\_SIMILARITY()}) in the SQL Cognitive Intelligence (CI) queries
(Figure~\ref{fig:example}) to extract semantic relationships between the relational values. The output of the semantic
functions is a numeric semantic score which is then used to order the
relational output (Figure~\ref{fig:example}). While the semantic score
can provide some insights towards understanding the results of semantic functions, it is still   
an abstraction, and does not provide detailed reasoning on the
\emph{causes} for the semantic score. Therefore, it is necessary
to understand the inner workings of the database embedding
model. 

db2Vec, like its counterparts in NLP such as word2Vec, generates
embedding vectors where each vector encodes a
\emph{distributed} representation of inferred semantics of the
corresponding text token~\cite{mikolov:nips13}. Every vector captures different attributes
of the inferred semantics, created in part by contributions of other
tokens within a \emph{context (neighborhood)}. The \emph{strength} of a
distributed relationship between tokens depends on how often these
tokens \emph{co-occur} within a context. Therefore, individual token
\textbf{occurrence} and \textbf{pair-wise co-occurrence counts} become key markers for
understanding influence of different tokens in the semantic model, and
can be used to provide deeper insights into the results of semantic functions.

\section{Interpretability Implementation and Usage}
\label{sec:impl}

AI-DB collects the token occurrence statistics and applies 
them to provide detailed reasoning into the ranked results of the semantic
SQL CI queries. Figure~\ref{fig:poverview} illustrates the three stages of integrating the 
interpretability capability into the AI-DB execution pipeline. The first stage(1) operates on 
the textified training document by extracting the required token statistics and storing the
information in a space-efficient manner (2). This information is then fetched at runtime for 
interpreting ranked results from a SQL CI query that invokes a semantic function (3). Given a ranked result list, the AI-DB interpretability
component aims to identify (1) columns from the base relational view, and (2) values in the base table, that have the most influence on the results.

\subsection{Token count Statistics}
From the training process outlined in the previous Section, one can see that
tokens with \texttt{NULL} values, as well as tokens associated with unique primary keys do not contribute to the
training and vector generation of other tokens. Since the tokens are
typed using the column names, it is possible to use the overall distribution of tokens across different columns 
to identify columns that play a significant role in the embedding process. To identify such columns, we have developed
two \emph{aggregate} metrics that use individual token counts to generate two per-column scores: Influence and Discriminatory scores.

\subsubsection{Influence Score}
The first metric, \emph{Influence Score}, captures the collective influence of a particular column on the overall training
as a measure of number of empty (\texttt{EMT}) tokens in that column. For a column $i$, it is calculated as a fraction of the number of \texttt{EMT} tokens to the
total number of tokens in column $i$:

{\small
\begin{equation}\label{infl}
Influence\_score(i) = 1.0 -\frac{Total(\#EMT\_Tokens(i))}{Total(\#Tokens(i))}
\end{equation}
}

The numerical ratio, varies from $1.0$ (most influence, with no \texttt{NULL} values in a column) to $0$ (all \texttt{NULL} values in a column).
The higher the influence score is for a column, the higher is the contribution of the values in that column to the trained model. 

\subsubsection{Discriminatory Score}
The \emph{Discriminatory Score} for a column $i$ captures the collective
ability of that column (i.e., aggregated over all tokens in that column) 
to semantically distinguish another entity in the base relational view. If a token 
appears multiple times in a column (i.e., it occurs in multiple rows or contexts), its final inferred
meaning is due to contributions from all of its neighboring tokens in the rows it where appears. Given a token, the larger the number 
of contributors is, lower is its ability to distinguish (or \emph{strongly relate}) to any of the tokens. For example, in the unique
primary key column, each token appears only once, and its inferred meaning is determined only by its neighbors in the associated row. On the
other extreme, if a particular column contains only one token, its inferred meaning is determined by all other tokens in the training document. In 
such case, the ability of the token to be dominantly related to another token is greatly diminished. For a column $i$, the Discriminatory score is
calculated as:

{\small
\begin{equation}\label{discr}
Discriminatory\_score(i) = \frac{Total(\#Unique\_Tokens(i))}{Total(\#Tokens(i))}
\end{equation}
}

The value of a column discriminatory score can vary from 1.0 (most discriminatory) to $\frac{1}{n}$ (least
discriminatory), where $n$ is the total number of tokens in that column. One can also compute the discriminatory score
per individual token, $tk_{i}$ as:

{\small
\begin{equation}\label{discr2}
Discriminatory\_score(tk_{i}) = 1.0- \frac{Total(\#Occurrences(tk_i))}{Total(\#Tokens(i))}
\end{equation}
}

The per-token discriminatory score value varies from $(n-1)/n$  (most discriminatory) to $0$ (least discriminatory).

These two scores encapsulate key characteristics of a training document (i.e., overall importance of columns and
unique tokens) and capture its impact on the trained model, and eventually on the quality of results from SQL CI queries. These
scores can be used to improve any issues with the training document (e.g., neglect the column with a low influence score (lots of \texttt{EMT} tokens), 
or with low discriminatory score (fewer unique values in a column). The column and per-token discriminatory scores can
also expose data skews in a column, and can be used to mitigate any
negative impacts (e.g., bias). In addition to these scores, AI-DB
provides detailed information about the numeric columns such as
minimum and maximum values as well as cluster information such as
cluster size, range, median, etc.

At runtime, for \emph{any} semantic function used in the SQL CI query, 
these scores can identify the most(least) influential columns and tokens 
\emph{solely} using the per-column token statistics for the model being used. 
Thus, the Influence and Discriminatory scores provide \emph{query-agnostic} \textbf{Model interpretability}.

\subsection{Token co-occurrence Statistics}
\label{sec:sketch}
The Influence and Discriminatory scores capture the overall importance of columns and
unique tokens during training. Token co-occurrence counts, on the other hand, focus
on capturing relationships between tokens associated with different types. 
As discussed in Section~\ref{sec:interpret} and Figure~\ref{fig:db2vec}, pair-wise
relationships form the \emph{atomic} building blocks for training the database embedding model. Thus,
pair-wise (bi-gram) co-occurrence counts become a reliable tool for understanding the causes
behind any db2Vec relationship at per-token granularity. 
\begin{table}
\caption{Unique co-occurring token pairs}
\centering
\begin{tabular}{ll}
\toprule
\multicolumn{1}{c}{Datasets} & \multicolumn{1}{c}{Unique} \\
 &  \multicolumn{1}{c}{Token Pairs}   \\
\cmidrule(lr){2-2}

Virginia                           & 13,481,992        \\
Fannie Mae                         & 203,974        \\
Airline                            & 5,902          \\
Mushroom                           & 5,462          \\ 
Credit Card Fraud                  & 7,094,716       \\
CA Toxicity                        & 13,413,076      \\
\bottomrule
\end{tabular}
\label{table:uniquepairs}
\end{table}

A typical database table may contain a fairly large number of entities, with potentially
significant number of unique values. Assuming $n$ unique tokens in a training document, the number of possible
pairs is $C(n,2) = \frac{n!}{(n-2)!2!} = \frac{n \times (n-1)}{2} \approx O(n^2)$. In other words, the size of the co-occurrence
matrix can be almost the square of the full vocabulary size! Table~\ref{table:uniquepairs} illustrates the number of 
co-occurring tokens for a representative set of training datasets.
For a larger value of $n$, this value is prohibitive and not all possible pairs can 
exist in a document. Therefore, solutions such as a naive array-based storage or a dictionary with full strings as keys, 
are both space inefficient and wasteful. 

The solution is to employ a probabilistic counting 
data structure, called Sketch~\cite{DBLP:journals/corr/PitelF15,journals/jal/CormodeM05,DBLP:journals/corr/abs-1902-10995}, 
to store and process the pair-wise co-occurrence counts.
Data sketches are probabilistic structures originally designed to record
aggregate counts of values in long-running (potentially infinite) streams (e.g., ad clicks)~\cite{gibbons:soda99, AMS96, cormode:tods}.
While sketches have been traditionally used for compression and
compactly recording tokens in streams for prespecified queries, it is being 
applied to other domains such as NLP~\cite{goyal-etal-2010-sketching} as well. Common
uses typically supported are total counts of tokens seen in a
stream or database, and a variety of algorithms, such the Count-Min (CM)
sketch have been developed for this purpose. AI-DB uses the CM sketch for storing and processing the token co-occurrence counts. 

Structurally, CM 
sketches are typically implemented as a two-dimensional table of with $h$ rows and $d$ columns. Each row corresponds to a 
different hash function over a range $d$. For every instance value to be recorded using $h$ independent
hash functions, distinct positions are calculated for each of the $h$ rows, and their values incremented.
At query time, for the token pair being queried, $h$ positions are first calculated, and then a count summarization is
performed across the rows using the minimum function over 
the positions determined by the hash functions, returning an approximation of the total count. The minimum function
helps to reduce the counting errors generated by the hash conflicts. In practice, sketches such as the CM sketch 
are shown to provide fairly accurate approximations of frequent item counts, 
but for low-frequency items the approximation errors can be significant~\cite{thomas:cell}. A sketch with $h$ hash tables with range $d$ each,
uses $O(h \times d)$ space, where $(h \times d) \ll n^2$. One could, conceivably, store the $O(n^2)$ co-occurrence matrix using sparse matrix 
serialization techniques, but the sparse approach, in the worst case, would still require space for storing $\frac{n \times (n-1)}{2}$ non-null values.

\begin{figure}[h]
\centering
\includegraphics[width=\linewidth]{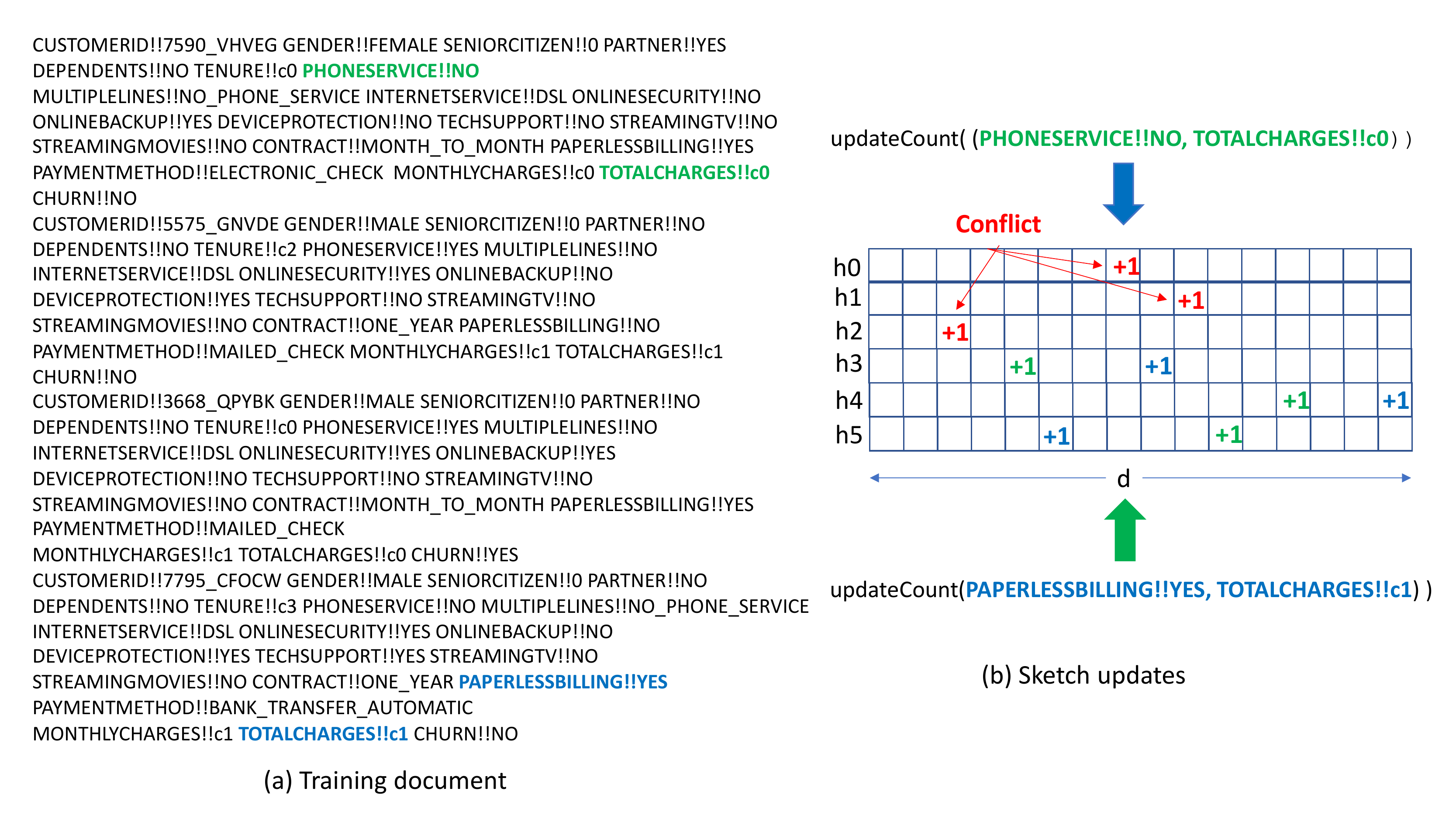}
\caption{Co-occurrence sketch count updates: (a) extracting token-pairs from the source training document, (b) updating the sketch per token-pair. The red values denote positions of conflicts during sketch updates.}
\label{fig:churn-sketch}
\end{figure}

In our implementation, a CM-based {\em co-occurrence sketch} is populated with the token-pair co-occurrence values
extracted from the textified document generated from a relational table (Figure~\ref{fig:poverview}(a)). For every "sentence"
in the training document (corresponds to a relational table row), each pair of tokens is selected (Figure~\ref{fig:churn-sketch}).
Then the CM co-occurrence sketch is updated as follows: (1) for the token-pair in consideration, $h$ positions are computed for each of the $h$ hash functions, and (2)
each value in the corresponding position is incremented by 1 (Figure~\ref{fig:poverview}(b)). The increment value represents the strength of the relationships between the tokens in a token-pair. Since database embedding assumes that all tokens associated with a relational row are equally related to each other, irrespective of the token-pair, every sketch increment has the same value. Tokens corresponding to unique primary keys (e.g., \texttt{CUSTOMERID}), appear
only once; hence any token pair containing such tokens is not stored in the co-occurrence sketch. It is possible that positions corresponding to 
different token pairs conflict (Figure~\ref{fig:churn-sketch}). However, the query-time minimum function has been designed to return 
a value that reduces the approximation error by returning the minimum value from the positions corresponding to a token-pair.
The implementation also uses a large hash function range, $d$. Larger the range is, the chances of
conflict reduce, decreasing the approximation error, and the co-occurrence sketch becomes increasingly sparse.

\begin{figure}[h]
\centering
\includegraphics[width=\linewidth]{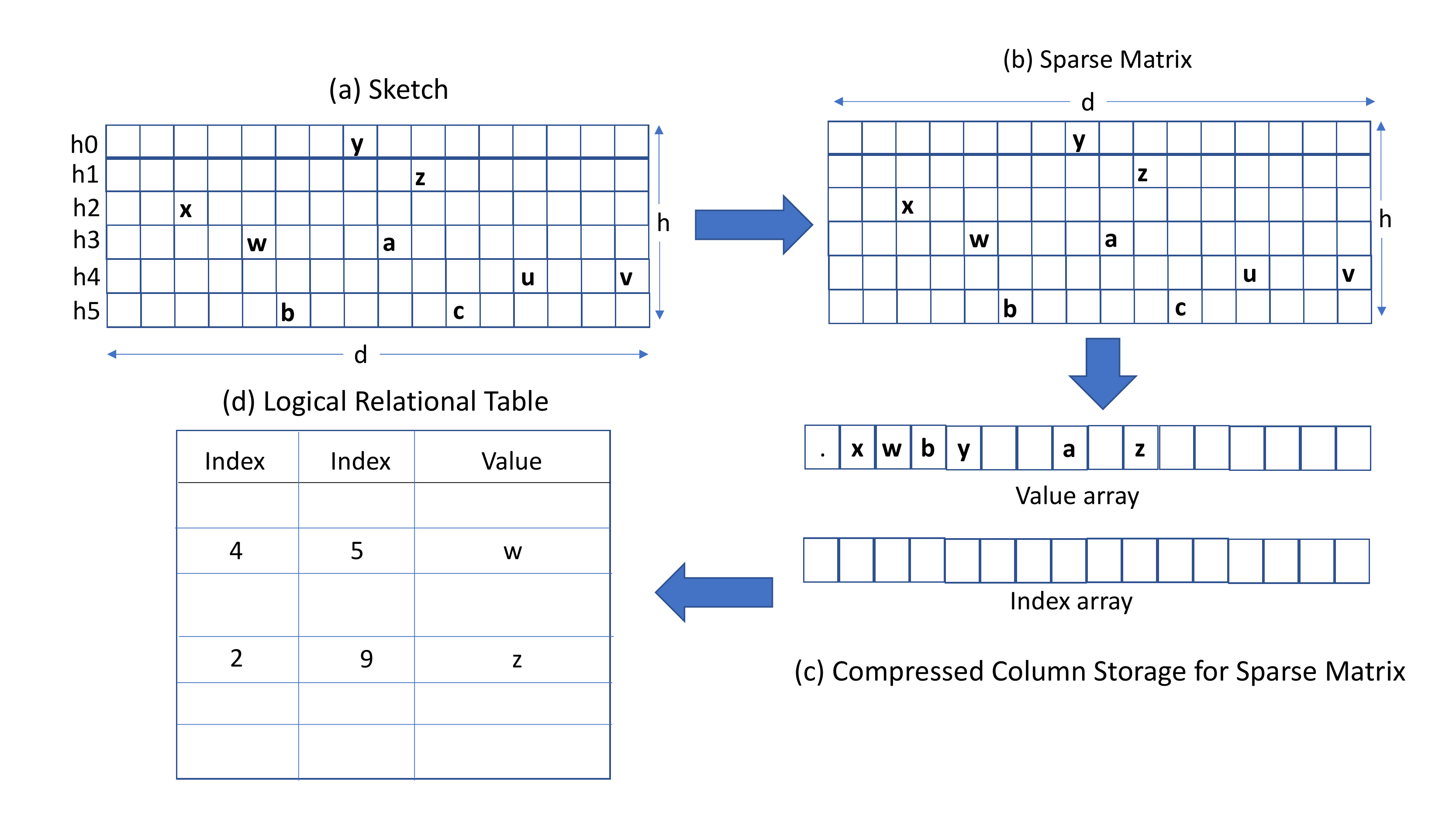}
\caption{Co-occurrence sketch as a sparse matrix: (1) co-occurrence sketch, (2) sparse matrix representation, (3) CSR
serialization, and (4) re-populating a logical relational table.}
\label{fig:sparsematrix}
\end{figure}

A benefit of using large sparse co-occurrence sketches is that they can be
viewed as sparse matrices as shown in
Figure~\ref{fig:sparsematrix}. As such, one can use any of several existing sparse storage representations,
e.g., Compressed Sparse Row (CSR) or Column (CSC)
formats~\cite{golub-book}, for efficiently storing the co-occurrence sketch. The
persistent size will be smaller than the in-memory representation and the
space reduction will be determined by the co-occurrence sketch sparsity. Specifically, a sparse
matrix serialization of a sketch with $h$ hash functions and range $d$, will store only $m$ non-null sketch values, where
$m < (h \times d) \ll n^2$.
The ability to increase the size while allowing the sparsity gives the best of both worlds:
accuracy in interpretability, while allowing for space and performance 
afforded by sparse matrix techniques. Once the co-occurrence sketch is populated, it
is serialized using the CSR sparse storage format that stores a matrix
in a dense row-major 
order using two arrays. The sparse 
matrix formulation enables resurrection of the co-occurrence sketch as a logical relational
table, where each row stores the row and column indices, along with  
the corresponding count value. At the time of interpreting results of
a SQL CI query (Figure~\ref{fig:poverview}), this relational
table can be queried to fetch the  
approximate counts of a token pair as follows: (1) For a token pair,
compute the position index $j$ for each of the $h$ hash
functions. (2) Use the hash 
(rows) index $i$, and the position index, $j$, to retrieve the
approximate count value.

An advantage of co-occurrence sketches is that they are sufficiently accurate
when the number of tokens is limited, and queries are made within those
limited numbers of inserted tokens.  If the universe of tokens used in
queries (such as tokens never seen before by the co-occurrence sketch) is much
larger than the co-occurrence sketch was designed for, the queries will produce
inaccurate counts.  This is common in database tables which have
columns with a large number of possible entries (such as account
numbers), and the total possible combinations, say in a database
transaction tracking to and from accounts, each with $m$ and $n$ possible
values would result in a large universe of $m \times n$ number of possible
co-occurring tokens; although the actual number of co-occurrences in
the database itself may be much smaller. Building sketches that store
counts for the large universe of all possible token co-occurrences is
prohibitive in terms of memory. To address this, we only
query the count sketch for tokens that exist in the database, by using
a Bloom filter-based~\cite{mitzen:book} shadow boolean sketch described in the following Section.


The space efficiency of a sketch can further be improved, although with a
small impact on accuracy, by {\em thresholding} small values in the
sketch and setting them to zero. This makes the sketch sparser, enabling further
reduction in the file size when the sketch is serialized using the CSR approach.

%

In addition to space efficiency, runtime performance 
is another major challenge facing the creation of sketches, 
especially as the associated table size grows. The sketch creation algorithm has
to track the co-occurrence in every row, and therefore the runtime
grows linearly with the size of the text file. Sketches are inherently
distributive in nature. As long as the size of the sketch, and the
hash functions for every row are the same, i.e., the sketches
are compatible, they can be simply added together without loss of
accuracy. Thus, speedup in processing can be achieved through
parallelizing the sketch creation process.  Each text file, can be split by rows into several
non-overlapping chunks, and separate count sketches are generated for
each chunk in parallel by separate threads.  Once the sketch
generation for the chunks is complete, the sketches generated by each
thread are merged together by adding the entries in the sketches into
a single top level sketch. 

All the necessary steps required for initializing, populating, storing, and using the co-occurrence count sketch
are built-in into the AI-DB processing flow (Figure~\ref{fig:poverview}) without using any external explainability tools
or packages, and requires no user intervention.

\subsection{Aggregate Statistics}
While the sketch enables the computation of co-occurrences, another additional capability is the gathering of statistical information about 
important relationships between pairs of variables.  In particular, for a given query token, the user may be interested in which other token 
has the highest co-occurrence with it, or interested in the median co-occurrences.  While the sketch does not store actual strings of co-occurrence
 pairs themselves, all pairs of strings can be computed with the addition of a small datastructure that only maintains the valid values for each column.  The additional data structure is a dictionary which has the column name as the key, and valid column categorical strings as the values.  Using this data structure, for any 
given query token, other tokens that have maximum co-occurrences can be found.

Consider a query token for which the maximum value needs to be computed. From the dictionary, it is paired with all valid values for other columns.  
The pairs are checked in the shadow sketch for existence. If the shadow sketch decides that the pair exists, then the co-occurrence sketch is queried. 
This is done for all pairs that are generated from the dictionary for the given token, and the maximum is computed.  
A shadow sketch, maintained for aggregate statistics, is implemented as a Bloom Filter which returns a boolean value to indicate 
whether a token pair generated by combining any two tokens is actually present in the sketch (and was present in the training set in the first place). 
This shadow sketch is only needed when computing aggregate statistics.

\subsection{Interpretability in Practice}
\label{sec:practice}

Let's describe how the AI-DB system uses the token count and co-occurrence 
statistics to interpret results of a SQL CI query. 
The AI-DB interpretability component (Figure~\ref{fig:poverview}) uses token 
occurrence statistics to provide query-agnostic column-oriented \textbf{model} interpretability and 
uses the pair-wise co-occurrence statistics to enable query-specific value-oriented \textbf{function} 
interpretability. The AI-DB function interpretability identifies
\emph{specific values} from the associated relational table that have the
most impact on the behavior of an semantic similarity function invoked by a CI query. 

\begin{figure}[htbp]
  \centering
  \includegraphics[width=\linewidth]{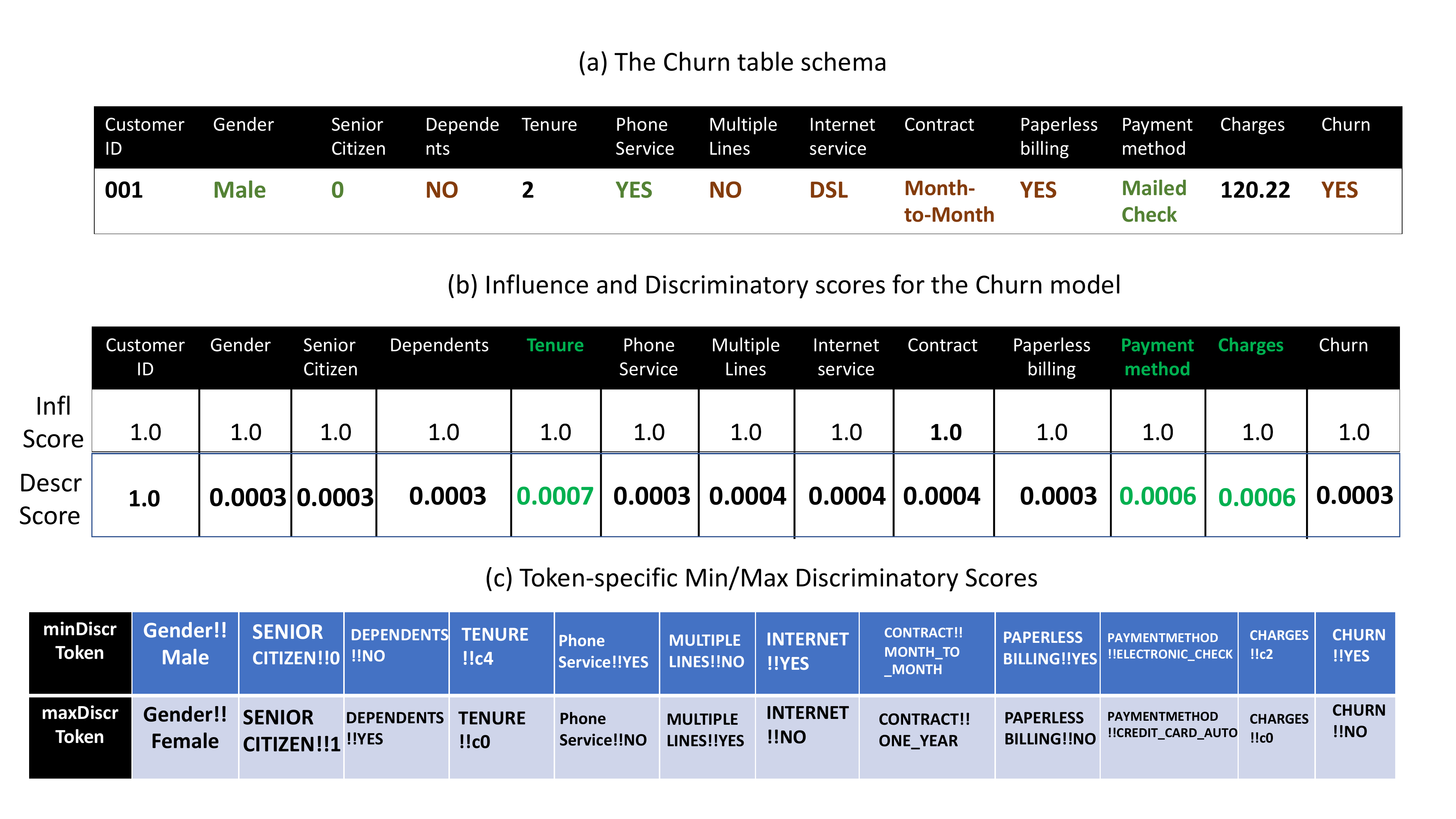}
  \caption{Influence and Discriminatory score for the Churn dataset. (a) Table schema with a sample row, (b) aggregated per-column influence
and discriminatory scores, and (c) per-column min and max discriminatory scores.}
  \label{fig:churn0}
\end{figure}

This section describes the AI-DB interpretability capability using the publicly available Telecom 
Churn dataset~\cite{churn-data}. The Churn dataset describes customer
characteristics of a hypothetical Telecom
company. Figure~\ref{fig:churn0}(a) presents the Churn table schema:
the table uses \texttt{CustomerID} as the primary key and contains
information about the customer account, including if the 
customer is considered to be churned. For training the db2Vec model, the \texttt{CustomerID} column
is defined as the key column, the \texttt{Charges} and \texttt{Tenure} columns are marked as the 
numerical columns, and the rest of the columns are marked as categorical. The values in the two 
numerical columns are clustered before training and replaced by tokens that represent clusters 
corresponding to the values. Figure~\ref{fig:churn0}(b) presents the per-column influence and 
discriminatory scores for the Churn table being trained. Since the table has no \texttt{NULL} 
values, all the columns have the highest influence score of 1.0. For the discriminator scores, the
\texttt{Tenure} column has the largest number of unique values (i.e., the number of clusters), resulting
in the second highest discriminatory score. The columns \texttt{Payment Method}, and \texttt{Charges} columns
have the next highest discriminatory scores. The remaining columns have very low discriminatory score
as the majority of columns only have binary values. Figure~\ref{fig:churn0}(c) presents per-column values
with minimum and maximum discriminator scores: the values with maximum occurrences have the least discriminatory score, 
and vice versa. This information enables the user to determine if the data is skewed, and which value is over-represented.
For example, from Figure~\ref{fig:churn0}(c), one can see that the
dataset is skewed towards young and male customers, with relatively
longer tenure (corresponding to the cluster \texttt{c4}), and paying
high charges (corresponding to the cluster \texttt{c2}) that 
are more likely to churn. One can use the data skew information to update the
raw dataset, or create a new view for training, or understand 
which columns and values impact the results.

\begin{figure}[htbp]
  \centering
  \includegraphics[width=\linewidth]{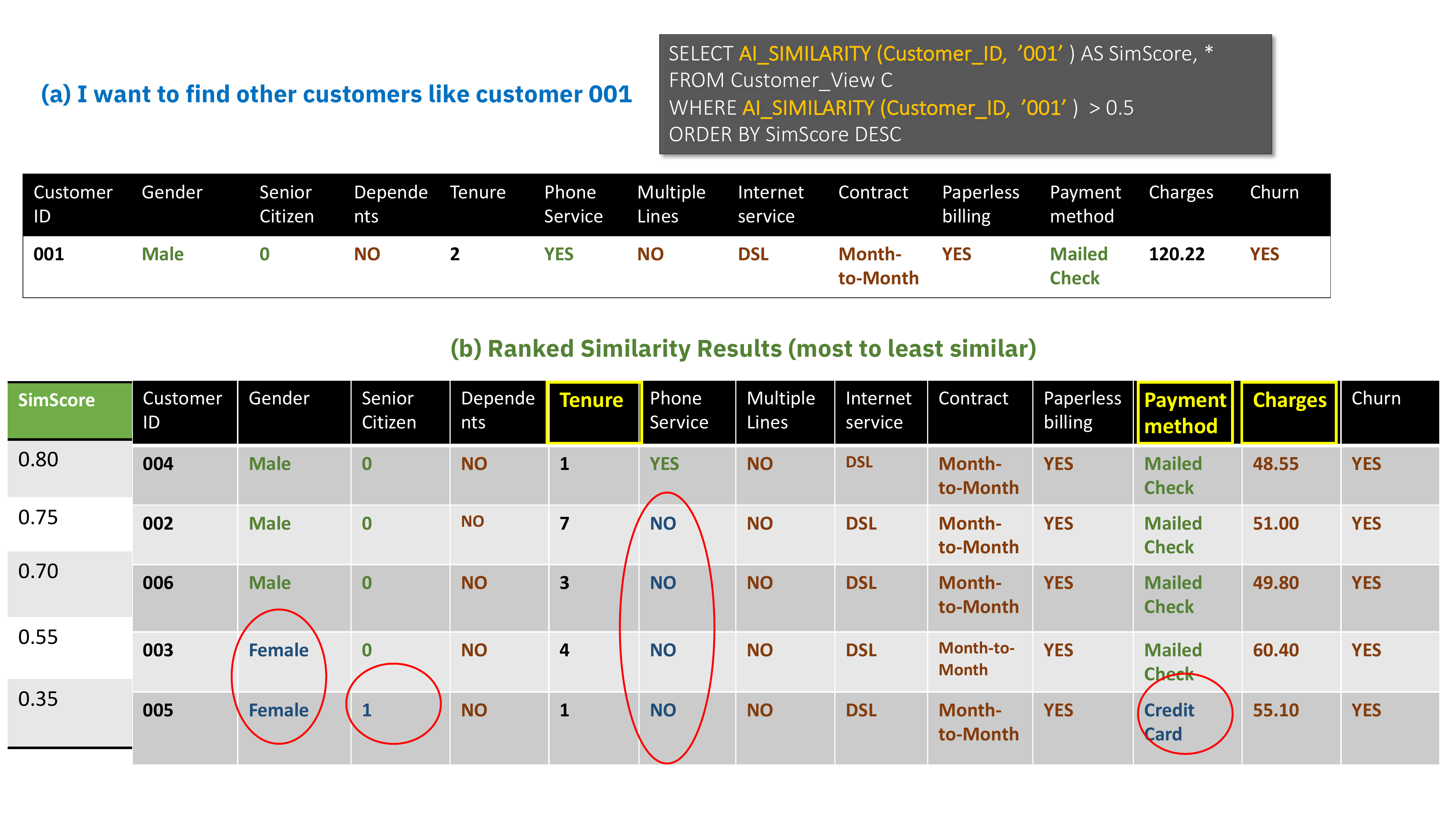}
  \caption{Invoking a primary key similarity query on the Churn dataset: (a) a sample semantic SQL query and the input row for customer \texttt{001}, and (b) ranked result rows with the most influential highlighted columns.}
  \label{fig:churn}
\end{figure}

AI-DB's built-in interpretability feature is illustrated by two exemplar pair-wise similarity queries.
The first type of query covers involves only primary key values and the second
involves internal (no-primary) values (the other value can be either
primary or internal value). The first example, Figure~\ref{fig:churn}
(a), presents a SQL CI query to find customer IDs similar to an
input customer ID (e.g., \texttt{001}). The SQL CI query returns the
output similarity score for the semantic function,
\texttt{AI\_SIMILARITY} as well as the values of other columns
(Figure~\ref{fig:churn}(b)). As the result shows, the most similar   
row has the highest similarity score (in this scenario, 
two primary key values are compared and the similarity score is  
the cosine similarity score between the corresponding vectors). Primary-key
based co-occurrence values are not stored in the co-occurrence sketch,
and as such only the model interpretability feature to understand 
results of the CI query. The
row corresponding to the most similar customer ID, \texttt{004},
matches in all categories with the row corresponding to customer
\texttt{001}. Numerical values corresponding to columns 
\texttt{Tenure} and \texttt{Charges} fall in the same range. Going 
down the ranked result list, it is observed that the rows have
increasing differences. Furthermore, based on the influence and
discriminatory scores, it is deduced that the \texttt{Tenure},
\texttt{Payment Method}, and \texttt{Charges} columns, have the most
impact on the result ordering. We can also observe that using the
per-token discriminator scores, the token,
\texttt{Tenure!!c0}, corresponding to the very small tenure values,
is the most impactful value in the result list. 

\begin{figure}[htbp]
  \centering
  \includegraphics[width=\linewidth]{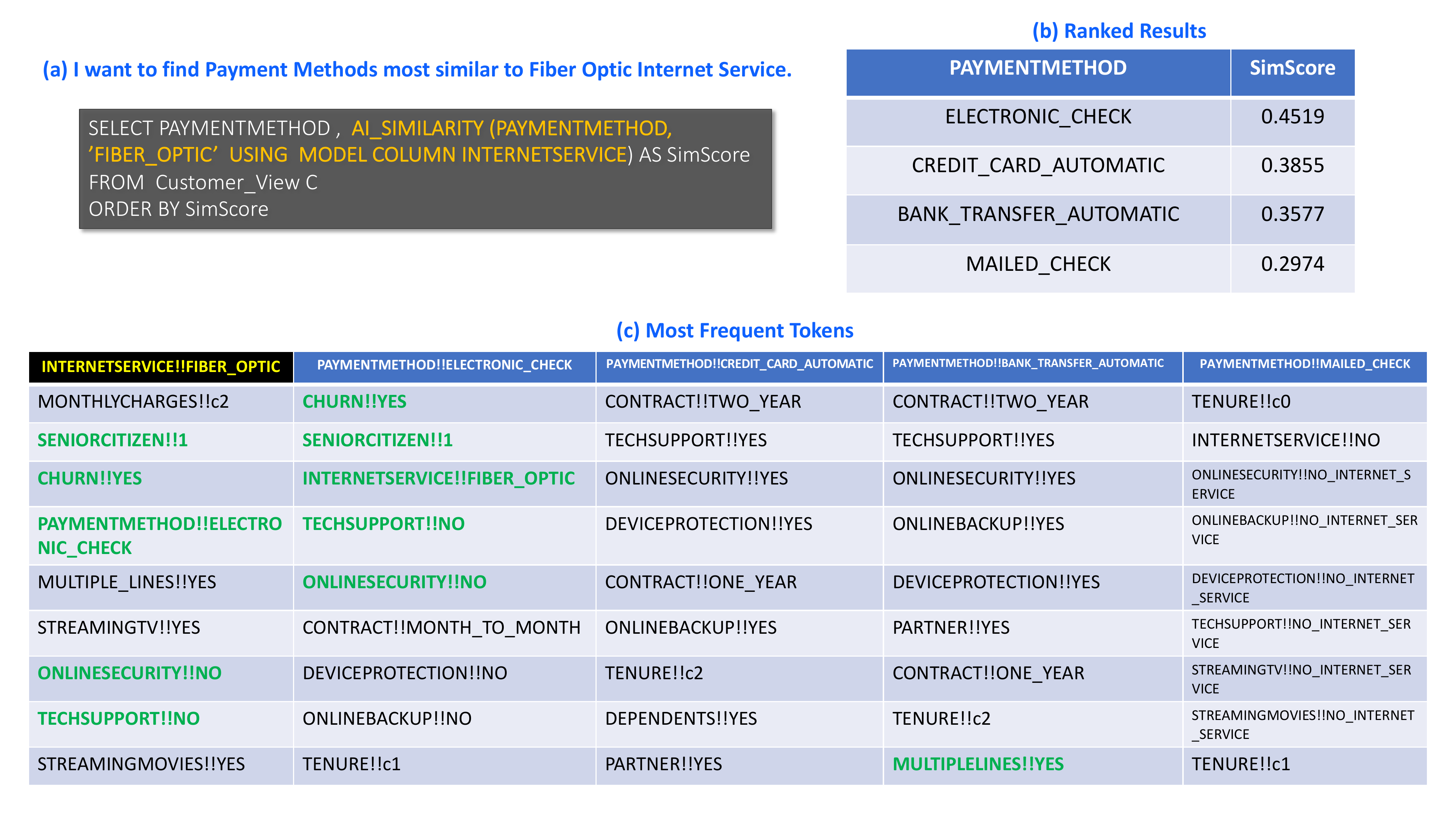}
  \caption{Invoking a similarity query on a non primary key value in the Churn dataset: (a) sample semantic SQL query, (b) sorted result, (c) 
most co-occurring tokens for the input and result values.}
  \label{fig:churn2}
\end{figure}

The second SQL CI query example, Figure~\ref{fig:churn2}(a), compares two non-primary
key values, associated with two different columns: it aims to find the
most related payment method to the \texttt{FIBER\_OPTIC} internet
service. Figure~\ref{fig:churn2}(b) presents the result list ordered
by the similarity score. To get insights into which values and columns
have the maximum impact on the sorted results, co-occurrence sketch is used. From this co-occurrence sketch, stored as an
auxiliary relational table, more frequent tokens of
associated with the key values in the query are obtained
(Figure~\ref{fig:churn2}(c)). From this list, one can observe that
\texttt{ELECTRONIC\_CHECK} is the most frequent payment method
associated with the \texttt{FIBER\_OPTIC} internet service. Furthermore,
\texttt{ELECTRONIC\_CHECK} and \texttt{FIBER\_OPTIC} share four
frequently occurring neighbors. Thus, using the co-occurrence counts
from the sketch, one can summarize that \texttt{ELECTRONIC\_CHECK} is
most related to \texttt{FIBER\_OPTIC} due to the strong \emph{primary} relationship
between the tokens, and the \emph{secondary} relationships generated by the
common neighbors (Section~\ref{sec:interpret}).

In practice, these two approaches are generalized to enable
interpretability for different semantic functions (e.g., dissimilarity, analogy, etc.) over
either primary key or internal values. For example, for a
dissimilarity query, the co-occurrence sketch can be used to identify
the \emph{least} co-occurring tokens to validate that output values
reported from  the query do not share any common
neighbors. Thus, using the similarity scores, along with model and
function interpretability capabilities, the AI-DB system is able
provide in-database support to provide key insights into the results
of the  SQL CI queries.

AI-DB's interpretability capability differs from the existing
interpretability/explainability approaches (Section~\ref{sec:related})
in many ways. First, techniques outlined in this Section are designed
for a self-supervised DL model that generates semantic embedded
vectors to be used for answering a variety of similarity based
queries. Although similar to vector embedding NLP models, the db2Vec 
model has key differences, e.g., support for non-text entities such as
numeric values. Our approach provides both query-agnostic
\emph{global} model interpretability, as well as, query-specific
\emph{local} function interpretability. The db2Vec training approach
provides hooks to extract key intrinsic information about
interpretability (i.e., db2Vec has been designed and implemented as a
self-describing model). We encode the relationships between the
trained entities using intrinsic co-occurrence properties of the
training data, and capture these relationships using a space-efficient
sketch data structure. The per-token discriminator score, as well as
the clustering information for the numeric columns, enables
identification of specific input \emph{values} that influence the
results of semantic queries. Finally, the entire interpretability pipeline
has been designed to be implemented within the context of a relational
database such that the required information can be extracted from an
input relational table, stored into an auxiliary relational table, and then used to interpret an SQL CI query
using the auxiliary relational table (Figure~\ref{fig:poverview}). Our 
interpretability approach does not use any external tools and requires no user 
intervention.

\section{Evaluation}
\label{sec:eval}

The evaluation of the quality of the proposed co-occurrence sketch for
functional interpretability as well as its runtime characteristics
such as space consumption and sketch generation performance, is detailed in this Section. For this evaluation, a set of datasets is chosen, 
10 pair-wise similarity queries for each dataset are run. The accuracy of the proposed sketch-based approach is evaluated for each query. 
An overview the selected datasets, description of the evaluation methodology, and experimental results are presented.  

\subsection{Evaluation Setup and Datasets}
\label{table:datasets}

The performance of the probabilistic sketch is validated by comparing
it's result accuracy to a {\em baseline approach} that gives exact results obtained using 
complete co-occurrence information in memory. The baseline system ingests
the complete training document as a Python Pandas dataframe, and
computes co-occurrence statistics over the entire training set.   The
{\em sketch approach} on the other hand uses the Spark \texttt{CountMinSketch}
Class~\cite{spark-sketch}, while the update and reads from this data structure is
performed with native Java. The code runs in Java native
multithreading mode for the larger data sets.  The experiments compare
the baseline and sketch approaches and score the comparison using the
NDCG metric. Both baseline and sketch approaches were implemented
in the Python-based AI-DB system (Figure~\ref{fig:poverview}) and the experiments reported in this
Section were run on x86/Linux machines. 

For the experimental evaluation, 6 different datasets
containing both real and synthetically generated data are used. An overview of
some dataset basic statistics is provided in Table \ref{tab:dsstructure}. In this table, the first two columns present 
the actual number of rows and columns in the source dataset. The third column presents the \emph{total} number of
non-primary key tokens and the last column reports the number of \emph{unique} non-primary key tokens in the corresponding textified (Section~\ref{sec:aidb}) training dataset. Note that number of unique tokens depends on the input data characteristics: a dataset dominated by either categorical
or numeric values tends to have fewer unique tokens after textification.

\begin{table}[htbp]
  \caption{Dataset structural characteristics: number of rows and columns along with counts of total and unique non-primary tokens}
  {\small
\centering
\begin{tabular}{lcccc}
  \toprule
  \multirow{2}{*}{Dataset} & \multicolumn{2}{c}{Table Information} & \# Non-primary & \# Non-primary  \\
  & Rows & Columns & Tokens & Unique Tokens \\ \cmidrule{2-3} \cmidrule{4-5} 
Virginia & 1,371,968 & 8 & 12,347,712 & 1,452,156        \\
Mushroom & 61,069 & 21 & 1,343,518 & 139        \\
Airline  & 129,880 & 23 & 3,117,120 & 115        \\
Fannie Mae & 11,232,359 & 20 & 235,879,539 & 1,475      \\
Credit Card & 14,820,425 & 15 & 237,126,800 & 111,752       \\
CA Toxicity & 1,248,836 & 88 & 111,146,404  & 92,567       \\
\bottomrule
\end{tabular}
}
\label{tab:dsstructure}
\end{table}

The Virginia dataset~\cite{virginia-data} is a publicly available
expenditure dataset from the State of Virginia for the 2015/2016
fiscal year. It lists details of every transaction such as vendor
name, corresponding state agency, which government fund was used etc,
by unique customers or vendors. Vendors can be individuals, and/or
private and public institutions, such as county agencies, school
districts, banks, businesses, etc. 

Freddie Mac and Fannie Mae are two Government agencies that buy
mortgages from lenders and hold them in portfolios or create sellable
mortgage backed securities. The Fannie-Mae dataset~\cite{fannie-mae-data} covers loan acquisition and performance data from
2007-2012 consisting of 30-year and less, fully amortizing, 
single-family, conventional fixed-rate mortgages.

The Credit Card transactional dataset~\cite{altman-credit-card} is a synthetically generated
dataset that contains transactions of about 2000 customers
with 40 fraudsters. Each row represents a unique
transaction with 15 distinct attributes such as user id, card id, time
based attributes, transaction amount, transaction type,  transaction
errors and finally merchant details like name, Merchant Category
Code(MCC), city, state and zip code.

The CA Toxicity dataset~\cite{ca-toxicity-data} reports results of
toxicity measurements for surface water performed at different
locations in the State of California. The Airline Passenger
Satisfaction dataset~\cite{airline-data} contains airline passenger
satisfaction survey covering different aspects of airline travel
experience (e.g., service, delays, distance, travel class, on-board
features, etc.) The UCI Mushroom dataset~\cite{mashroom-data} contains
information on 23 species of mushrooms such as physical, visual and
olfactory features of each specimen as well as habitat and growth
pattern information and whether the specimen is edible or not.

\subsection{Evaluation Methodology}

For each dataset, ten different pair-wise similarity queries were
performed. These queries covered inter- and intra-column
query patterns over non-primary keys (Section~\ref{sec:practice}). Given an input dataset, it is first
converted to a text training document used as the input
to generate probabilistic co-occurrence sketch as well as a precise,
but space consuming, co-occurrence data structure. Using these two
data structures, the top co-occurring tokens are computed for the input
query parameter and the result values from the corresponding training text file.  
Co-occurrences for each of the query and
results are limited to the top five columns of interest in the
corresponding relational table and are presented in sorted order. The \emph{exact} results
serve as a {\em baseline} for comparison and validation of the
\emph{approximate} results from the probabilistic co-occurrence
sketch.

The exact and approximate lists are then compared using the Normalized
Discounted Cumulative Gain (NDCG)~\cite{ndcgref} metric. NDCG is a
widely used metric to evaluate search results in order of
relevance. NDCG compares not only a sequence of results, but the
sorted order of relevance in which they are presented as well.  The
NDCG score used in this paper differs from the traditional computation
in two important ways. First, there are no repeated values, 
and second, an additional penalty parameter is added that reduces the NDCG score
in proportion to the difference in expected and observed
positions. The higher the NDCG score is, the higher is the matching, both in
the values in the list, and its order. An NDCG score of 1.0 means
that both lists contain the same entries in the same order.

\begin{figure}[h]
\centering
\includegraphics[width=\linewidth]{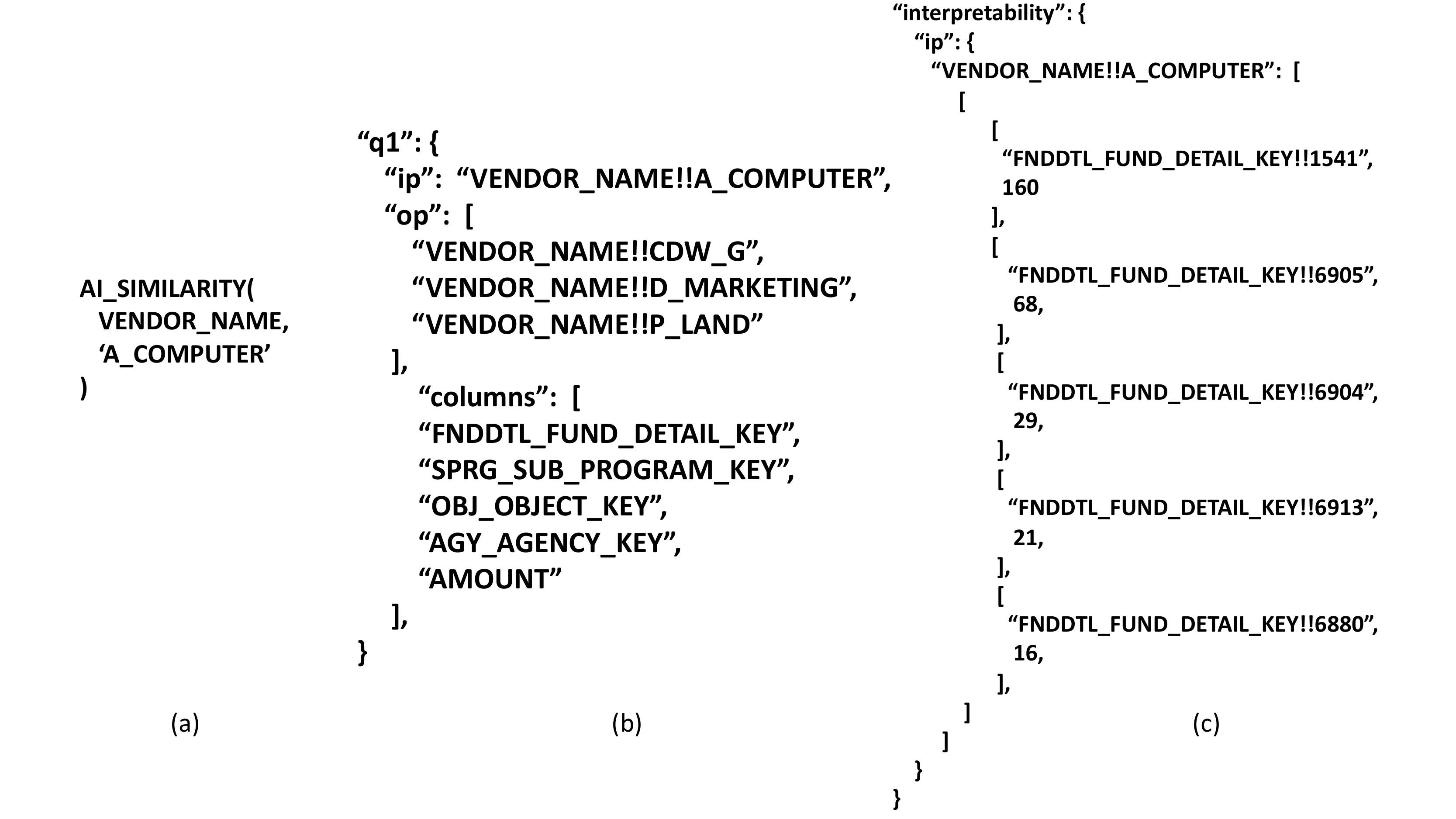}
\caption{Example of one of the queries for the Virginia dataset (a), JSON representation of the results (b), and co-occurrence results (c)}
\label{fig:querysample}
\end{figure}

To illustrate this evaluation approach, an example for the
Virginia dataset is used: Figure~\ref{fig:querysample}(a) shows an intra-column query comparing a specified
vendor, \texttt{A\_COMPUTER} with other
vendors. Figure~\ref{fig:querysample}(b) shows the corresponding JSON
representation. The keyword {\em q1}
denotes the intra-column similarity query, {\em ip} denotes the token
on which the similarity query was performed, {\em op} denotes the tokens obtained
as results from AI-DB for the query, and  {\em columns} shows the
table columns of interest for the input parameter.  For the query {\em
  q1}, five columns were specified as of interest to the user.

Figure~\ref{fig:querysample}(c) gives a small exemplary snippet (due to
paper space limitations) of the interpretability co-occurrence results
for the query in Figure~\ref{fig:querysample}(a).  The
snippet shows the co-occurrence results (tokens and their co-occurrence counts) for one column, \newline \texttt{
  FNDDTL\_FUND\_DETAIL\_KEY}, in sorted order for the input parameter
\texttt{VENDOR\_NAME!!A\_COMPUTER}. As discussed in
Figure~\ref{fig:churn2}, functional interpretability requires the list
of most co-occurring tokens both for the input and result
values (e.g., \texttt{VENDOR\_NAME!!CDW\_G}). Therefore, the full JSON file stores similar key-value 
results for each input and result values of each similarity query
(Figure~\ref{fig:querysample}(a)), for each of the top five 
selected columns. Once the JSON files are generated,  JSON files for the
same set of query inputs as well as output results are scored for the baseline
and the sketch scenarios. The scoring
is based on NDCG comparison of the sorted lists for each input query,
and result as described earlier, and an average of the scores
for that experiment are presented.   

\subsection{Evaluation Results}

In this section, the behavior of the co-occurrence sketch along three dimensions is evaluated: relative
accuracy for the queries as defined by the NDCG metric, space utilization, and multithreaded scalability.

\begin{flushleft}
\textbf{Evaluating Space Utilization:}
\end{flushleft}

Table~\ref{tab:dsstructure} gives insight into the structural aspects
of the textified training files that are used to generate the co-occurrence sketch.
Parameters of interest such as number of database table rows,  number of
total tokens, and number of unique tokens (vocabulary) are presented.
 These parameters provide an insight into the challenges in creating a
 co-occurrence sketch for the dataset.  While the number of 
lines in the textified file decides the total time for data parsing
and  creating the co-occurrence sketch, the number of unique tokens and the related number
of co-occurrences of tokens impacts the size of co-occurrence sketch
(as well as the likelihood of hash conflicts in a less than sufficient 
sized sketch).  

\begin{table*}[h]
\caption{Space savings with co-occurrence sketch (using 5 hash
  functions with NDCG of 0.95 or higher). Sketch range $d$ in millions (M). Sparse file stored in the CSR format.}
\centering
\begin{tabular}{lcccccccc}  
\toprule
\multirow{2}{*}{Dataset}  & Estimated Space(MB)  &  Sparse File Size(MB)  & Co-occurrence Sketch & Space \\
  & for Token Pairs & for Co-occurrence Sketch & Range (M) $d$  & Saving \\ \cmidrule{2-5}
Virginia & 448 & 58.2 & 8 & 7.82 \\ 
Mushroom & 0.46 & 0.074 & 0.1 & 6.21      \\ 
Airline  & 0.33 & 0.052 & 0.1  & 6.34       \\ 
Fannie Mae & 17.94  & 2.8 & 4  & 6.4       \\ 
Credit Card & 424.84 & 64.78 & 16  & 6.57       \\ 
CA Toxicity & 640.72 & 71.8 & 32  & 8.92       \\ 
\bottomrule
\end{tabular}%
\label{tab:dsarea}
\end{table*}

Table~\ref{tab:dsarea} illustrates the space savings provided by the sketch as
compared to the estimated space for storing the co-occurrence values
as a dictionary in Java.  There are several potential ways to track
co-occurrences.  First, a textified file (which is an entire database
table in text form) may be stored in memory and co-occurrences calculated on demand,
but this is inefficient and prohibitive from a memory
standpoint. Another possibility is a dictionary based data structure,
where each co-occurring pair of tokens in the data set is maintained a
key with the total count for that co-occurring token as the value.  In
any data set, the number of pairs of tokens can be quite large, making
a dictionary approach also inefficient for storage in memory, although
it would be definitely preferable to storing the entire database
table. 

The sketch on the other hand, does not store the actual string for the
co-occurring pair. Instead, it uses a hash on the string to identify
locations in the arrays to maintain the counts.  In hash based data
structures, there is a trade-off between accuracy due to has hash
conflicts and size of the structure.  The results in
Table~\ref{tab:dsarea} compare the size of a sketch that provides an
NDCG accuracy of at least 0.95 when compared to an estimate of the conservative base
case of storing the data structure as a dictionary.

As can be seen from the results, a gain of
5x-10x in memory area was achieved with the sketch as compared to a
dictionary data structure for an NDCG accuracy score of 0.95 or
greater.  The memory savings are significantly even higher for lower
but acceptable NDCG accuracy scores. Since the use of the sketch is
for interpretability of results already produced by the semantic functions, small
variations in sorted order have lower impact on the interpretability
quality. 

\begin{flushleft}
\textbf{Evaluating Sketch Accuracy:}
\end{flushleft}

As with any approach based on hashing, increasing the size of the data
structure reduces hash conflicts, and therefore improves accuracy.
Count-min structures with multiple rows improve the accuracy of
hashing for smaller sized data structures while allowing some amount
of hashing conflict within a given location.  By using a Count-min
sketch, we keep the accuracy higher even with smaller sizes for the
rows, and allowing for some hash conflict.  Table~\ref{tab:acc} gives
the accuracy sensitivity of a sketch for the Virginia, Fannie Mae and
CA Toxicity datasets. 

\begin{table*}[htbp]
\caption{Accuracy vs. space utilization for Virginia, Fannie Mae, and CA Toxicity datasets. The co-occurrence sketch is using 5 hash functions ($h=5$). Sketch range $d$ in millions (M).}  
\centering
\begin{tabular}{ccccccc}
  \toprule
\multicolumn{2}{c}{Virginia} & \multicolumn{2}{c}{Fannie Mae} & \multicolumn{2}{c}{CA Toxicity} \\
Sketch Range (M) & NDCG & Sketch Range (M) & NDCG & Sketch Range (M) & NDCG \\
  $d$ & Accuracy &  $d$ & Accuracy  &  $d$ & Accuracy \\ \cmidrule{1-2} \cmidrule{3-4} \cmidrule{5-6}
1          &  0.56  & 0.03            & 0.58 & 0.125  & 0.54\\
2            & 0.56  & 0.06            & 0.66 & 0.5  & 0.56\\
3           & 0.62  & 0.125           & 0.69 & 1  & 0.67 \\
4          & 0.64 & 0.25         & 0.76 & 2  & 0.76 \\ 
5          & 0.79 & 0.5          & 0.83 & 4  & 0.83\\ 
6          & 0.78 & 1          & 0.86 &  6 & 0.90\\ 
7          & 0.81 & 2          & 0.92 & 8 &  0.92\\ 
8         & 0.99 & 3         &  0.94 & 16  & 0.93\\ 
9          & 0.99 & 4          & 0.95 & 24 &  0.94\\ 
10       & 0.99  & 5       &   0.99 & 32 &  0.97\\
\bottomrule
\end{tabular}
\label{tab:acc}
\end{table*}

As described earlier with regard to accuracy with space trade-off,
the size of the \emph{in-memory} co-occurrence sketch can be grown to increase accuracy, and due to the
sparsity of the structure, the overall file space required can still be managed.
Table~\ref{tab:dssparsity} gives the percentage of the sketch that
stores the zero value for NDCG accuracy targets of 0.95 and 0.99.
In Table~\ref{tab:dssparsity},
the Sketch Range column reflects an effort to increase the accuracy
of the probabilistic co-occurrence sketch to 0.95 and above by increasing the number of entries in the hash range $d$ (values reported in units of millions (M)).
The CSR sparse matrix file size is compared against the total area of the in-memory sketch for
each of these target accuracies. 
Due to the sparsity of the sketch data structure, it is possible to
compress the space using a sparse matrix representation while targeting
high accuracy from the sketch. With the increased accuracy targets of 0.99,
significantly larger sketches are required (in a few datasets double), 
but overall the CSR size increases by a much smaller amount. The increased
sparsity helps to improve accuracy by managing
the overhead of storing these sketch tables in a relational database table during runtime.

\begin{table*}[htbp]
\caption{Sparsity with sketch sizes for NDCG values: 0.95 and 0.99. Sketch range $d$ in millions (M). Sparse file stored in the CSR format.}
\centering
\begin{tabular}{lcccccccc}  
\toprule
\multirow{3}{*}{Dataset}&\multicolumn{4}{c}{NDCG=0.95} & \multicolumn{4}{c}{NDCG=0.99}\\
                        &  Sketch      & Full Sketch   & Sparse File &  Sparsity&   Sketch & Full Sketch   & Sparse File & Sparsity\\
                        & Range(M) $d$ & Size (MB)   & Size (MB) &  \% Zero Values         & Range(M) $d$ & Size (MB)   &  Size (MB) &  \% Zero Values \\ \cmidrule{3-5} \cmidrule{7-9}
Virginia                &8              & 160         & 58.2               & 63       & 16            &    320      & 78.4   & 75\\
Mushroom                &0.1            & 1           & 0.074              & 92       & 0.1         &      2      & 0.18 & 91\\
Airline                 &0.1            & 1           & 0.052              & 94       & 0.1         &      2      & 0.14 & 93\\
Fannie Mae              &4              & 80          & 2.8                 & 96       & 5             & 100      & 3.1   & 97\\
Credit Card             &16             & 320         & 64.78               & 79       & 32            & 640      & 77.0  & 88\\
CA Toxicity             &32             & 640         & 71.8                & 89       & 64           & 1280     & 104   & 92\\
\bottomrule
\end{tabular}%
\label{tab:dssparsity}
\end{table*}

\begin{flushleft}
\textbf{Evaluating Runtime Scalability:}
\end{flushleft}

The sketch data structure is inherently distributive. Sketches can be
created from mutually exclusive subgroups of database rows or lines
of a textified file of a dataset. The individual sketches generated
from different groups of lines can be then merged without locking, simply by adding
together the values at each array position into a single sketch, as
long as each individual table has the same dimension and each row
has the same hash function across the different sketches.  If each of
the individual sketches is a two-dimensional vector of the same
dimension and each row is also associated with the same hash function
across the table vectors, the resulting sketch is a vector addition of
the individual vectors. 

\begin{figure}[h]
\centering
\includegraphics[width=\linewidth]{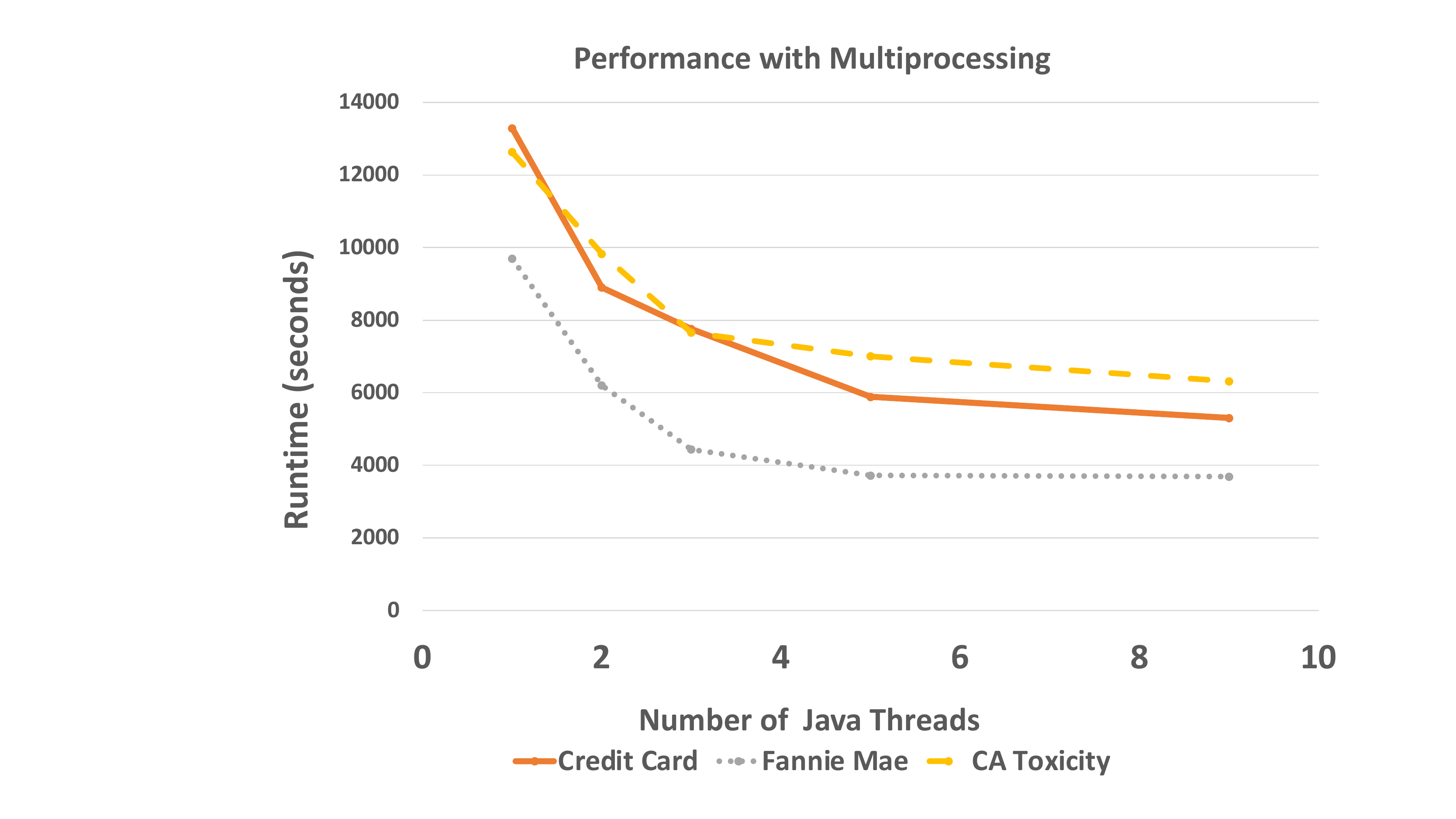}
\caption{Scaling the co-occurrence sketch generation using multiple Java threads. Performance scaling reduces for the number of threads greater than 5.}
\label{fig:multiscale}
\end{figure}

Table~\ref{tab:runtime} provides the amount of time taken to build the
sketch.  For the relatively smaller data sets, e.g., Virginia, single threaded runs
were used. For larger datasets such as Fannie Mae, Credit Card, and CA
Toxicity, Table~\ref{tab:runtime} reports the sketch-building runtimes. The number of threads
represents the thread count, after which multiprocessing scaling improvements become
insignificant. Figure~\ref{fig:multiscale} gives the improvements for the
Credit Card data set (one the larger data sets) with
multithreading. In this particular test case, the scaling improvements
tend to level off after about 5 threads, due to the overhead in
creating a full sized sketch with each thread, and the costs of
merging as each thread requires it's own copy of a full sized sketch.
Experimental data demonstrated that the co-occurrence data structure      
provides accurate results with space-efficient footprint and scalable build performance.

\begin{table}[htbp]
\caption{Sketch build runtime (in milliseconds) along with the corresponding number of threads}  
\centering
\begin{tabular}{lcc}  
\toprule
\multirow{2}{*}{Dataset}                  & Build Time   & \multirow{2}{*}{\# Threads}   \\
                                          & milliseconds & \\ \cmidrule{2-3}
Virginia                          & 94,137  &  1 \\ 
Mushroom                         & 21,688  &  1 \\ 
Airline                           & 29,063  &  1 \\ 
Fannie Mae                        & 5,587,096   & 5  \\ 
Credit Card                       & 2,952,080   & 6  \\ 
CA Toxicity                       & 7,419,687   & 5  \\
\bottomrule
\end{tabular}
\label{tab:runtime}
\end{table}

\section{Conclusions}
\label{sec:concl}

A novel in-database interpretability system for
providing detailed insights on the results of semantic SQL (CI)
queries is presented. This capability has been designed to support a
self-supervised vector embedding model that can be trained over very
large database tables. A novel design that uses a
probabilistic Count-Min sketch data structure to store the data co-occurrence
information using a sparse-matrix data serialization format
(CSR) is introduced. The sketch is materialized at runtime as a relational table which is queried for
analyzing results of semantic functions. Experimental evaluation
on a set of large and distinct datasets demonstrated that the sketch
data structure provides a space-efficient and scalable approach to
provide accurate interpretability results. As a future work, techniques for reducing Sketch conflicts using the
Frequency-aware Count-Min (FCM) approach~\cite{thomas:cell}, as well as enhancing the
core interpretability capabilities to support new semantic queries will be investigated.

\section{Acknowledgments}

The authors thank Jose Neves, Matt Tong, and Wei Zhang for their detailed comments on the earlier versions of this paper.

\balance

\bibliographystyle{ACM-Reference-Format}
\bibliography{ai-related, refs}


\begin{thebibliography}{58}


\ifx \showCODEN    \undefined \def \showCODEN     #1{\unskip}     \fi
\ifx \showDOI      \undefined \def \showDOI       #1{#1}\fi
\ifx \showISBNx    \undefined \def \showISBNx     #1{\unskip}     \fi
\ifx \showISBNxiii \undefined \def \showISBNxiii  #1{\unskip}     \fi
\ifx \showISSN     \undefined \def \showISSN      #1{\unskip}     \fi
\ifx \showLCCN     \undefined \def \showLCCN      #1{\unskip}     \fi
\ifx \shownote     \undefined \def \shownote      #1{#1}          \fi
\ifx \showarticletitle \undefined \def \showarticletitle #1{#1}   \fi
\ifx \showURL      \undefined \def \showURL       {\relax}        \fi
\providecommand\bibfield[2]{#2}
\providecommand\bibinfo[2]{#2}
\providecommand\natexlab[1]{#1}
\providecommand\showeprint[2][]{arXiv:#2}

\bibitem[\protect\citeauthoryear{Allen and Hospedales}{Allen and
  Hospedales}{2019}]%
        {allen:analogy}
\bibfield{author}{\bibinfo{person}{Carl Allen} {and}
  \bibinfo{person}{Timothy~M. Hospedales}.} \bibinfo{year}{2019}\natexlab{}.
\newblock \showarticletitle{Analogies Explained: Towards Understanding Word
  Embeddings}.
\newblock \bibinfo{journal}{\emph{CoRR}}  \bibinfo{volume}{abs/1901.09813}
  (\bibinfo{year}{2019}).
\newblock
\showeprint[arXiv]{1901.09813}
\urldef\tempurl%
\url{http://arxiv.org/abs/1901.09813}
\showURL{%
\tempurl}


\bibitem[\protect\citeauthoryear{Alon, Matias, and Szegedy}{Alon
  et~al\mbox{.}}{1996}]%
        {AMS96}
\bibfield{author}{\bibinfo{person}{Noga Alon}, \bibinfo{person}{Yossi Matias},
  {and} \bibinfo{person}{Mario Szegedy}.} \bibinfo{year}{1996}\natexlab{}.
\newblock \showarticletitle{The Space Complexity of Approximating the Frequency
  Moments}. In \bibinfo{booktitle}{\emph{ACM Symposium on Theory of
  Computing}}. \bibinfo{pages}{20--29}.
\newblock


\bibitem[\protect\citeauthoryear{Altman}{Altman}{2019}]%
        {altman-credit-card}
\bibfield{author}{\bibinfo{person}{Erik Altman}.}
  \bibinfo{year}{2019}\natexlab{}.
\newblock \bibinfo{title}{Credit Card Transactions Dataset}.
\newblock
\newblock
\urldef\tempurl%
\url{https://www.kaggle.com/datasets/ealtman2019/credit-card-transactions}
\showURL{%
\tempurl}


\bibitem[\protect\citeauthoryear{{Apache Spark}}{{Apache Spark}}{2022}]%
        {spark-sketch}
\bibfield{author}{\bibinfo{person}{{Apache Spark}}.}
  \bibinfo{year}{2022}\natexlab{}.
\newblock \bibinfo{title}{Online documentation}.
\newblock
\newblock
\urldef\tempurl%
\url{https://spark.apache.org/docs/preview/api/java/org/apache/spark/util/sketch/CountMinSketch.html}
\showURL{%
\tempurl}


\bibitem[\protect\citeauthoryear{Arrieta, Rodr{\'{\i}}guez, Ser, Bennetot,
  Tabik, Barbado, Garc{\'{\i}}a, Gil{-}Lopez, Molina, Benjamins, Chatila, and
  Herrera}{Arrieta et~al\mbox{.}}{2019}]%
        {arrieta:xai}
\bibfield{author}{\bibinfo{person}{Alejandro~Barredo Arrieta},
  \bibinfo{person}{Natalia~D{\'{\i}}az Rodr{\'{\i}}guez},
  \bibinfo{person}{Javier~Del Ser}, \bibinfo{person}{Adrien Bennetot},
  \bibinfo{person}{Siham Tabik}, \bibinfo{person}{Alberto Barbado},
  \bibinfo{person}{Salvador Garc{\'{\i}}a}, \bibinfo{person}{Sergio
  Gil{-}Lopez}, \bibinfo{person}{Daniel Molina}, \bibinfo{person}{Richard
  Benjamins}, \bibinfo{person}{Raja Chatila}, {and} \bibinfo{person}{Francisco
  Herrera}.} \bibinfo{year}{2019}\natexlab{}.
\newblock \showarticletitle{Explainable Artificial Intelligence {(XAI):}
  Concepts, Taxonomies, Opportunities and Challenges toward Responsible {AI}}.
\newblock \bibinfo{journal}{\emph{CoRR}}  \bibinfo{volume}{abs/1910.10045}
  (\bibinfo{year}{2019}).
\newblock
\showeprint[arXiv]{1910.10045}
\urldef\tempurl%
\url{http://arxiv.org/abs/1910.10045}
\showURL{%
\tempurl}


\bibitem[\protect\citeauthoryear{Bahdanau, Cho, and Bengio}{Bahdanau
  et~al\mbox{.}}{2014}]%
        {bahdanau:arxiv}
\bibfield{author}{\bibinfo{person}{Dzmitry Bahdanau},
  \bibinfo{person}{Kyunghyun Cho}, {and} \bibinfo{person}{Yoshua Bengio}.}
  \bibinfo{year}{2014}\natexlab{}.
\newblock \bibinfo{title}{Neural Machine Translation by Jointly Learning to
  Align and Translate}.
\newblock
\newblock
\urldef\tempurl%
\url{https://doi.org/10.48550/ARXIV.1409.0473}
\showDOI{\tempurl}


\bibitem[\protect\citeauthoryear{Belinkov and Glass}{Belinkov and
  Glass}{2018}]%
        {belinkov:survey}
\bibfield{author}{\bibinfo{person}{Yonatan Belinkov} {and}
  \bibinfo{person}{James~R. Glass}.} \bibinfo{year}{2018}\natexlab{}.
\newblock \showarticletitle{Analysis Methods in Neural Language Processing: {A}
  Survey}.
\newblock \bibinfo{journal}{\emph{CoRR}}  \bibinfo{volume}{abs/1812.08951}
  (\bibinfo{year}{2018}).
\newblock
\showeprint[arXiv]{1812.08951}
\urldef\tempurl%
\url{http://arxiv.org/abs/1812.08951}
\showURL{%
\tempurl}


\bibitem[\protect\citeauthoryear{Bibal, Cardon, Alfter, Wilkens, Wang,
  Fran{\c{c}}ois, and Watrin}{Bibal et~al\mbox{.}}{2022}]%
        {bibal-etal-2022-attention}
\bibfield{author}{\bibinfo{person}{Adrien Bibal}, \bibinfo{person}{R{\'e}mi
  Cardon}, \bibinfo{person}{David Alfter}, \bibinfo{person}{Rodrigo Wilkens},
  \bibinfo{person}{Xiaoou Wang}, \bibinfo{person}{Thomas Fran{\c{c}}ois}, {and}
  \bibinfo{person}{Patrick Watrin}.} \bibinfo{year}{2022}\natexlab{}.
\newblock \showarticletitle{Is Attention Explanation? An Introduction to the
  Debate}. In \bibinfo{booktitle}{\emph{Proceedings of the 60th Annual Meeting
  of the Association for Computational Linguistics (Volume 1: Long Papers)}}.
  \bibinfo{publisher}{Association for Computational Linguistics},
  \bibinfo{address}{Dublin, Ireland}, \bibinfo{pages}{3889--3900}.
\newblock
\urldef\tempurl%
\url{https://doi.org/10.18653/v1/2022.acl-long.269}
\showDOI{\tempurl}


\bibitem[\protect\citeauthoryear{Bordawekar}{Bordawekar}{2018}]%
        {spark-cogdb}
\bibfield{author}{\bibinfo{person}{Rajesh Bordawekar}.}
  \bibinfo{year}{2018}\natexlab{}.
\newblock \bibinfo{title}{{Cognitive Database: An Apache Spark-Based AI-Enabled
  Relational Database System}}.
\newblock \bibinfo{howpublished}{Spark Data+AI Summit}.
\newblock


\bibitem[\protect\citeauthoryear{Bordawekar and Shmueli}{Bordawekar and
  Shmueli}{2016}]%
        {bordawekar:corr-abs-1603-07185}
\bibfield{author}{\bibinfo{person}{Rajesh Bordawekar} {and}
  \bibinfo{person}{Oded Shmueli}.} \bibinfo{year}{2016}\natexlab{}.
\newblock \showarticletitle{Enabling Cognitive Intelligence Queries in
  Relational Databases using Low-dimensional Word Embeddings}.
\newblock \bibinfo{journal}{\emph{CoRR}}  \bibinfo{volume}{abs/1603.07185}
  (\bibinfo{year}{2016}).
\newblock
\urldef\tempurl%
\url{http://arxiv.org/abs/1603.07185}
\showURL{%
\tempurl}


\bibitem[\protect\citeauthoryear{Bordawekar and Shmueli}{Bordawekar and
  Shmueli}{2017}]%
        {Bordawekar:deem17}
\bibfield{author}{\bibinfo{person}{Rajesh Bordawekar} {and}
  \bibinfo{person}{Oded Shmueli}.} \bibinfo{year}{2017}\natexlab{}.
\newblock \showarticletitle{Using Word Embedding to Enable Semantic Queries in
  Relational Databases}. In \bibinfo{booktitle}{\emph{Proceedings of the 1st
  Workshop on Data Management for End-to-End Machine Learning}} (Chicago, IL,
  USA) \emph{(\bibinfo{series}{DEEM'17})}. \bibinfo{publisher}{ACM},
  \bibinfo{address}{New York, NY, USA}, Article \bibinfo{articleno}{5},
  \bibinfo{numpages}{4}~pages.
\newblock
\showISBNx{978-1-4503-5026-6}
\urldef\tempurl%
\url{https://doi.org/10.1145/3076246.3076251}
\showDOI{\tempurl}


\bibitem[\protect\citeauthoryear{Celli}{Celli}{2021}]%
        {lex2vec}
\bibfield{author}{\bibinfo{person}{Fabio Celli}.}
  \bibinfo{year}{2021}\natexlab{}.
\newblock \showarticletitle{Lex2vec: making Explainable Word Embedding via
  Distant Supervision}.
\newblock \bibinfo{journal}{\emph{CoRR}}  \bibinfo{volume}{abs/2103.02269}
  (\bibinfo{year}{2021}).
\newblock
\showeprint[arXiv]{2103.02269}
\urldef\tempurl%
\url{https://arxiv.org/abs/2103.02269}
\showURL{%
\tempurl}


\bibitem[\protect\citeauthoryear{Church and Hanks}{Church and Hanks}{1990}]%
        {church-hanks-1990-word}
\bibfield{author}{\bibinfo{person}{Kenneth~Ward Church} {and}
  \bibinfo{person}{Patrick Hanks}.} \bibinfo{year}{1990}\natexlab{}.
\newblock \showarticletitle{Word Association Norms, Mutual Information, and
  Lexicography}.
\newblock \bibinfo{journal}{\emph{Computational Linguistics}}
  \bibinfo{volume}{16}, \bibinfo{number}{1} (\bibinfo{year}{1990}),
  \bibinfo{pages}{22--29}.
\newblock
\urldef\tempurl%
\url{https://aclanthology.org/J90-1003}
\showURL{%
\tempurl}


\bibitem[\protect\citeauthoryear{{Commonwealth of Virginia}}{{Commonwealth of
  Virginia}}{2016}]%
        {virginia-data}
\bibfield{author}{\bibinfo{person}{{Commonwealth of Virginia}}.}
  \bibinfo{year}{2016}\natexlab{}.
\newblock \bibinfo{title}{State of virginia 2016 expenditures}.
\newblock
\newblock
\urldef\tempurl%
\url{https://www.datapoint.apa.virginia.gov}
\showURL{%
\tempurl}


\bibitem[\protect\citeauthoryear{Cormode and Muthukrishnan}{Cormode and
  Muthukrishnan}{2005a}]%
        {journals/jal/CormodeM05}
\bibfield{author}{\bibinfo{person}{Graham Cormode} {and} \bibinfo{person}{S.
  Muthukrishnan}.} \bibinfo{year}{2005}\natexlab{a}.
\newblock \showarticletitle{An improved data stream summary: the count-min
  sketch and its applications.}
\newblock \bibinfo{journal}{\emph{J. Algorithms}} \bibinfo{volume}{55},
  \bibinfo{number}{1} (\bibinfo{year}{2005}), \bibinfo{pages}{58--75}.
\newblock
\urldef\tempurl%
\url{http://dblp.uni-trier.de/db/journals/jal/jal55.html#CormodeM05}
\showURL{%
\tempurl}


\bibitem[\protect\citeauthoryear{Cormode and Muthukrishnan}{Cormode and
  Muthukrishnan}{2005b}]%
        {cormode:tods}
\bibfield{author}{\bibinfo{person}{Graham Cormode} {and} \bibinfo{person}{S.
  Muthukrishnan}.} \bibinfo{year}{2005}\natexlab{b}.
\newblock \showarticletitle{What's Hot and What's Not: Tracking Most Frequent
  Items Dynamically}.
\newblock \bibinfo{journal}{\emph{ACM Transactions on Database Systems}}
  \bibinfo{volume}{30}, \bibinfo{number}{1} (\bibinfo{date}{March}
  \bibinfo{year}{2005}), \bibinfo{pages}{249--278}.
\newblock


\bibitem[\protect\citeauthoryear{Danilevsky, Qian, Aharonov, Katsis, Kawas, and
  Sen}{Danilevsky et~al\mbox{.}}{2020}]%
        {DBLP:journals/corr/abs-2010-00711}
\bibfield{author}{\bibinfo{person}{Marina Danilevsky}, \bibinfo{person}{Kun
  Qian}, \bibinfo{person}{Ranit Aharonov}, \bibinfo{person}{Yannis Katsis},
  \bibinfo{person}{Ban Kawas}, {and} \bibinfo{person}{Prithviraj Sen}.}
  \bibinfo{year}{2020}\natexlab{}.
\newblock \showarticletitle{A Survey of the State of Explainable {AI} for
  Natural Language Processing}.
\newblock \bibinfo{journal}{\emph{CoRR}}  \bibinfo{volume}{abs/2010.00711}
  (\bibinfo{year}{2020}).
\newblock
\showeprint[arXiv]{2010.00711}
\urldef\tempurl%
\url{https://arxiv.org/abs/2010.00711}
\showURL{%
\tempurl}


\bibitem[\protect\citeauthoryear{Date}{Date}{1982}]%
        {DBLP:journals/sigmod/Date82}
\bibfield{author}{\bibinfo{person}{C.~J. Date}.}
  \bibinfo{year}{1982}\natexlab{}.
\newblock \showarticletitle{A Formal Definition of the relational Model}.
\newblock \bibinfo{journal}{\emph{{SIGMOD} Rec.}} \bibinfo{volume}{13},
  \bibinfo{number}{1} (\bibinfo{year}{1982}), \bibinfo{pages}{18--29}.
\newblock
\urldef\tempurl%
\url{https://doi.org/10.1145/984514.984515}
\showDOI{\tempurl}


\bibitem[\protect\citeauthoryear{Du, Liu, and Hu}{Du et~al\mbox{.}}{2020}]%
        {du:cacm-intp}
\bibfield{author}{\bibinfo{person}{Mengnan Du}, \bibinfo{person}{Ninghao Liu},
  {and} \bibinfo{person}{Xia Hu}.} \bibinfo{year}{2020}\natexlab{}.
\newblock \showarticletitle{Techniques for interpretable machine learning}.
\newblock \bibinfo{journal}{\emph{Commun. ACM}} \bibinfo{volume}{63},
  \bibinfo{number}{1} (\bibinfo{date}{January} \bibinfo{year}{2020}).
\newblock


\bibitem[\protect\citeauthoryear{Face}{Face}{2023}]%
        {hugface}
\bibfield{author}{\bibinfo{person}{Hugging Face}.}
  \bibinfo{year}{2023}\natexlab{}.
\newblock \bibinfo{title}{Hugging Face Models}.
\newblock
\newblock
\urldef\tempurl%
\url{https://huggingface.co/models}
\showURL{%
\tempurl}


\bibitem[\protect\citeauthoryear{{Fannie Mae}}{{Fannie Mae}}{2022}]%
        {fannie-mae-data}
\bibfield{author}{\bibinfo{person}{{Fannie Mae}}.}
  \bibinfo{year}{2022}\natexlab{}.
\newblock \bibinfo{title}{{Fannie Mae Single-Family Loan Performance Data}}.
\newblock
\newblock
\urldef\tempurl%
\url{https://capitalmarkets.fanniemae.com/credit-risk-transfer/single-family-credit-risk-transfer/fannie-mae-single-family-loan-performance-data}
\showURL{%
\tempurl}


\bibitem[\protect\citeauthoryear{Gibbons and Matias}{Gibbons and
  Matias}{1999}]%
        {gibbons:soda99}
\bibfield{author}{\bibinfo{person}{Phillip~B. Gibbons} {and}
  \bibinfo{person}{Yossi Matias}.} \bibinfo{year}{1999}\natexlab{}.
\newblock \showarticletitle{Synopsis Data Structures for Massive Data Sets}. In
  \bibinfo{booktitle}{\emph{Proceedings of Symposium on Discrete Algorithms}}.
\newblock


\bibitem[\protect\citeauthoryear{Gilpin, Bau, Yuan, Bajwa, Specter, and
  Kagal}{Gilpin et~al\mbox{.}}{2018}]%
        {DBLP:journals/corr/abs-1806-00069}
\bibfield{author}{\bibinfo{person}{Leilani~H. Gilpin}, \bibinfo{person}{David
  Bau}, \bibinfo{person}{Ben~Z. Yuan}, \bibinfo{person}{Ayesha Bajwa},
  \bibinfo{person}{Michael~A. Specter}, {and} \bibinfo{person}{Lalana Kagal}.}
  \bibinfo{year}{2018}\natexlab{}.
\newblock \showarticletitle{Explaining Explanations: An Approach to Evaluating
  Interpretability of Machine Learning}.
\newblock \bibinfo{journal}{\emph{CoRR}}  \bibinfo{volume}{abs/1806.00069}
  (\bibinfo{year}{2018}).
\newblock
\showeprint[arXiv]{1806.00069}
\urldef\tempurl%
\url{http://arxiv.org/abs/1806.00069}
\showURL{%
\tempurl}


\bibitem[\protect\citeauthoryear{Golub and Van~Loan}{Golub and
  Van~Loan}{1996}]%
        {golub-book}
\bibfield{author}{\bibinfo{person}{Gene Golub} {and} \bibinfo{person}{Charles
  Van~Loan}.} \bibinfo{year}{1996}\natexlab{}.
\newblock \bibinfo{booktitle}{\emph{Matrix Computations}
  (\bibinfo{edition}{3rd} ed.)}.
\newblock \bibinfo{publisher}{Johns Hopkins}, \bibinfo{address}{Baltimore,
  Maryland}.
\newblock


\bibitem[\protect\citeauthoryear{Goyal, Jagarlamudi, Daum{\'e}~III, and
  Venkatasubramanian}{Goyal et~al\mbox{.}}{2010}]%
        {goyal-etal-2010-sketching}
\bibfield{author}{\bibinfo{person}{Amit Goyal}, \bibinfo{person}{Jagadeesh
  Jagarlamudi}, \bibinfo{person}{Hal Daum{\'e}~III}, {and}
  \bibinfo{person}{Suresh Venkatasubramanian}.}
  \bibinfo{year}{2010}\natexlab{}.
\newblock \showarticletitle{Sketching Techniques for Large Scale {NLP}}. In
  \bibinfo{booktitle}{\emph{Proceedings of the {NAACL} {HLT} 2010 Sixth Web as
  Corpus Workshop}}. \bibinfo{publisher}{Association for Computational
  Linguistics}, \bibinfo{address}{NAACL-HLT, Los Angeles},
  \bibinfo{pages}{17--25}.
\newblock
\urldef\tempurl%
\url{https://aclanthology.org/W10-1503}
\showURL{%
\tempurl}


\bibitem[\protect\citeauthoryear{Ho}{Ho}{2022}]%
        {ho:dl-complexity}
\bibfield{author}{\bibinfo{person}{Tin~Kam Ho}.}
  \bibinfo{year}{2022}\natexlab{}.
\newblock \showarticletitle{Complexity of Representations in Deep Learning}.
\newblock  (\bibinfo{year}{2022}).
\newblock
\urldef\tempurl%
\url{https://doi.org/10.48550/ARXIV.2209.00525}
\showDOI{\tempurl}


\bibitem[\protect\citeauthoryear{IBM}{IBM}{2020}]%
        {churn-data}
\bibfield{author}{\bibinfo{person}{IBM}.} \bibinfo{year}{2020}\natexlab{}.
\newblock \bibinfo{title}{Predict Customer Churn using Watson Machine Learning
  and Jupyter Notebooks on Cloud Pak for Data}.
\newblock
\newblock
\urldef\tempurl%
\url{https://github.com/IBM/telco-customer-churn-on-icp4d}
\showURL{%
\tempurl}
\newblock
\shownote{Apache-2.0 License.}


\bibitem[\protect\citeauthoryear{{IBM}}{{IBM}}{2022}]%
        {ibm-zADE}
\bibfield{author}{\bibinfo{person}{{IBM}}.} \bibinfo{year}{2022}\natexlab{}.
\newblock \bibinfo{title}{{IBM Z Artificial Intelligence Data Embedding
  Library}}.
\newblock \bibinfo{howpublished}{{IBM z/OS 2.5.0 Online Documentation}}.
\newblock


\bibitem[\protect\citeauthoryear{{IBM Db2 13 Project Team}}{{IBM Db2 13 Project
  Team}}{2022}]%
        {db2-sdi}
\bibfield{author}{\bibinfo{person}{{IBM Db2 13 Project Team}}.}
  \bibinfo{year}{2022}\natexlab{}.
\newblock \bibinfo{title}{{IBM Db2 13 for z/OS and More}}.
\newblock \bibinfo{howpublished}{IBM Redbook}.
\newblock
\urldef\tempurl%
\url{https://www.ibm.com/products/db2-for-zos}
\showURL{%
\tempurl}


\bibitem[\protect\citeauthoryear{Jain and Wallace}{Jain and Wallace}{2019}]%
        {DBLP:journals/corr/abs-1902-10186}
\bibfield{author}{\bibinfo{person}{Sarthak Jain} {and}
  \bibinfo{person}{Byron~C. Wallace}.} \bibinfo{year}{2019}\natexlab{}.
\newblock \showarticletitle{Attention is not Explanation}.
\newblock \bibinfo{journal}{\emph{CoRR}}  \bibinfo{volume}{abs/1902.10186}
  (\bibinfo{year}{2019}).
\newblock
\showeprint[arXiv]{1902.10186}
\urldef\tempurl%
\url{http://arxiv.org/abs/1902.10186}
\showURL{%
\tempurl}


\bibitem[\protect\citeauthoryear{J\"{a}rvelin and
  Kek\"{a}l\"{a}inen}{J\"{a}rvelin and Kek\"{a}l\"{a}inen}{2002}]%
        {ndcgref}
\bibfield{author}{\bibinfo{person}{Kalervo J\"{a}rvelin} {and}
  \bibinfo{person}{Jaana Kek\"{a}l\"{a}inen}.} \bibinfo{year}{2002}\natexlab{}.
\newblock \showarticletitle{Cumulated Gain-Based Evaluation of IR Techniques}.
\newblock \bibinfo{journal}{\emph{ACM Trans. Inf. Syst.}} \bibinfo{volume}{20},
  \bibinfo{number}{4} (\bibinfo{date}{oct} \bibinfo{year}{2002}),
  \bibinfo{pages}{422–446}.
\newblock
\showISSN{1046-8188}
\urldef\tempurl%
\url{https://doi.org/10.1145/582415.582418}
\showDOI{\tempurl}


\bibitem[\protect\citeauthoryear{{Karl Pearson}}{{Karl Pearson}}{1901}]%
        {pca}
\bibfield{author}{\bibinfo{person}{{Karl Pearson}}.}
  \bibinfo{year}{1901}\natexlab{}.
\newblock \showarticletitle{{LIII. On lines and planes of closest fit to
  systems of points in space}}.
\newblock \bibinfo{journal}{\emph{The London, Edinburgh, and Dublin
  Philosophical Magazine and Journal of Science}} \bibinfo{volume}{2},
  \bibinfo{number}{11} (\bibinfo{year}{1901}), \bibinfo{pages}{559--572}.
\newblock
\urldef\tempurl%
\url{https://doi.org/10.1080/14786440109462720}
\showDOI{\tempurl}
\showeprint{https://doi.org/10.1080/14786440109462720}


\bibitem[\protect\citeauthoryear{Klein}{Klein}{2020}]%
        {airline-data}
\bibfield{author}{\bibinfo{person}{T.~J. Klein}.}
  \bibinfo{year}{2020}\natexlab{}.
\newblock \bibinfo{title}{Airline Passenger Satisfaction dataset}.
\newblock
\newblock
\urldef\tempurl%
\url{https://www.kaggle.com/datasets/teejmahal20/airline-passenger-satisfaction}
\showURL{%
\tempurl}


\bibitem[\protect\citeauthoryear{Lundberg and Lee}{Lundberg and Lee}{2017}]%
        {lundberg:shap}
\bibfield{author}{\bibinfo{person}{Scott~M. Lundberg} {and}
  \bibinfo{person}{Su{-}In Lee}.} \bibinfo{year}{2017}\natexlab{}.
\newblock \showarticletitle{A unified approach to interpreting model
  predictions}.
\newblock \bibinfo{journal}{\emph{CoRR}}  \bibinfo{volume}{abs/1705.07874}
  (\bibinfo{year}{2017}).
\newblock
\showeprint[arXiv]{1705.07874}
\urldef\tempurl%
\url{http://arxiv.org/abs/1705.07874}
\showURL{%
\tempurl}


\bibitem[\protect\citeauthoryear{Madsen, Reddy, and Chandar}{Madsen
  et~al\mbox{.}}{2021}]%
        {madsen:survey}
\bibfield{author}{\bibinfo{person}{Andreas Madsen}, \bibinfo{person}{Siva
  Reddy}, {and} \bibinfo{person}{Sarath Chandar}.}
  \bibinfo{year}{2021}\natexlab{}.
\newblock \showarticletitle{Post-hoc Interpretability for Neural {NLP:} {A}
  Survey}.
\newblock \bibinfo{journal}{\emph{CoRR}}  \bibinfo{volume}{abs/2108.04840}
  (\bibinfo{year}{2021}).
\newblock
\showeprint[arXiv]{2108.04840}
\urldef\tempurl%
\url{https://arxiv.org/abs/2108.04840}
\showURL{%
\tempurl}


\bibitem[\protect\citeauthoryear{Marcinkevics and Vogt}{Marcinkevics and
  Vogt}{2020}]%
        {ricards:zoo}
\bibfield{author}{\bibinfo{person}{Ricards Marcinkevics} {and}
  \bibinfo{person}{Julia~E. Vogt}.} \bibinfo{year}{2020}\natexlab{}.
\newblock \showarticletitle{Interpretability and Explainability: {A} Machine
  Learning Zoo Mini-tour}.
\newblock \bibinfo{journal}{\emph{CoRR}}  \bibinfo{volume}{abs/2012.01805}
  (\bibinfo{year}{2020}).
\newblock
\showeprint[arXiv]{2012.01805}
\urldef\tempurl%
\url{https://arxiv.org/abs/2012.01805}
\showURL{%
\tempurl}


\bibitem[\protect\citeauthoryear{Mikolov, Sutskever, Chen, Corrado, and
  Dean}{Mikolov et~al\mbox{.}}{2013}]%
        {mikolov:nips13}
\bibfield{author}{\bibinfo{person}{Tomas Mikolov}, \bibinfo{person}{Ilya
  Sutskever}, \bibinfo{person}{Kai Chen}, \bibinfo{person}{Gregory~S. Corrado},
  {and} \bibinfo{person}{Jeffrey Dean}.} \bibinfo{year}{2013}\natexlab{}.
\newblock \showarticletitle{Distributed Representations of Words and Phrases
  and their Compositionality}. In \bibinfo{booktitle}{\emph{27th Annual
  Conference on Neural Information Processing Systems 2013.}}
  \bibinfo{pages}{3111--3119}.
\newblock


\bibitem[\protect\citeauthoryear{Mitzenmacher and Upfal}{Mitzenmacher and
  Upfal}{2005}]%
        {mitzen:book}
\bibfield{author}{\bibinfo{person}{Michael Mitzenmacher} {and}
  \bibinfo{person}{Eli Upfal}.} \bibinfo{year}{2005}\natexlab{}.
\newblock \bibinfo{booktitle}{\emph{Probability and Computing: Randomized
  Algorithms and Probabilistic Analysis}}.
\newblock \bibinfo{publisher}{Cambridge University Press}.
\newblock
\showISBNx{0-521-83540-2}


\bibitem[\protect\citeauthoryear{Molnar}{Molnar}{2022}]%
        {molnar2022}
\bibfield{author}{\bibinfo{person}{Christoph Molnar}.}
  \bibinfo{year}{2022}\natexlab{}.
\newblock \bibinfo{title}{Interpretable Machine Learning}.
\newblock
\newblock
\urldef\tempurl%
\url{https://christophm.github.io/interpretable-ml-book}
\showURL{%
\tempurl}


\bibitem[\protect\citeauthoryear{Neelakantan, Xu, Puri, Radford, Han, Tworek,
  Yuan, Tezak, Kim, Hallacy, Heidecke, Shyam, Power, Nekoul, Sastry, Krueger,
  Schnurr, Such, Hsu, Thompson, Khan, Sherbakov, Jang, Welinder, and
  Weng}{Neelakantan et~al\mbox{.}}{2022}]%
        {openai-text}
\bibfield{author}{\bibinfo{person}{Arvind Neelakantan}, \bibinfo{person}{Tao
  Xu}, \bibinfo{person}{Raul Puri}, \bibinfo{person}{Alec Radford},
  \bibinfo{person}{Jesse~Michael Han}, \bibinfo{person}{Jerry Tworek},
  \bibinfo{person}{Qiming Yuan}, \bibinfo{person}{Nikolas Tezak},
  \bibinfo{person}{Jong~Wook Kim}, \bibinfo{person}{Chris Hallacy},
  \bibinfo{person}{Johannes Heidecke}, \bibinfo{person}{Pranav Shyam},
  \bibinfo{person}{Boris Power}, \bibinfo{person}{Tyna~Eloundou Nekoul},
  \bibinfo{person}{Girish Sastry}, \bibinfo{person}{Gretchen Krueger},
  \bibinfo{person}{David Schnurr}, \bibinfo{person}{Felipe~Petroski Such},
  \bibinfo{person}{Kenny Hsu}, \bibinfo{person}{Madeleine Thompson},
  \bibinfo{person}{Tabarak Khan}, \bibinfo{person}{Toki Sherbakov},
  \bibinfo{person}{Joanne Jang}, \bibinfo{person}{Peter Welinder}, {and}
  \bibinfo{person}{Lilian Weng}.} \bibinfo{year}{2022}\natexlab{}.
\newblock \showarticletitle{Text and Code Embeddings by Contrastive
  Pre-Training}.
\newblock \bibinfo{journal}{\emph{CoRR}}  \bibinfo{volume}{abs/2201.10005}
  (\bibinfo{year}{2022}).
\newblock
\showeprint[arXiv]{2201.10005}
\urldef\tempurl%
\url{https://arxiv.org/abs/2201.10005}
\showURL{%
\tempurl}


\bibitem[\protect\citeauthoryear{Neves, Bordawekar, and Tzortzatos}{Neves
  et~al\mbox{.}}{2019}]%
        {neves:aidb19}
\bibfield{author}{\bibinfo{person}{Jose Neves}, \bibinfo{person}{Rajesh
  Bordawekar}, {and} \bibinfo{person}{Elpida Tzortzatos}.}
  \bibinfo{year}{2019}\natexlab{}.
\newblock \showarticletitle{{Demonstrating Semantic SQL Queries over Relational
  Data using the AI-Powered Database}}. In
  \bibinfo{booktitle}{\emph{Proceedings of the 1st International Workshop on
  Applied AI for Database Systems and Applications (AIDB’19)}}.
\newblock


\bibitem[\protect\citeauthoryear{{Nvidia}}{{Nvidia}}{2023}]%
        {nvidia-llm}
\bibfield{author}{\bibinfo{person}{{Nvidia}}.} \bibinfo{year}{2023}\natexlab{}.
\newblock \bibinfo{title}{{Explore NVIDIA NeMo LLM Service}}.
\newblock
\newblock
\urldef\tempurl%
\url{https://www.nvidia.com/en-us/deep-learning-ai/solutions/large-language-models}
\showURL{%
\tempurl}


\bibitem[\protect\citeauthoryear{Park, Bak, and Oh}{Park et~al\mbox{.}}{2017}]%
        {park-etal-2017-rotated}
\bibfield{author}{\bibinfo{person}{Sungjoon Park}, \bibinfo{person}{JinYeong
  Bak}, {and} \bibinfo{person}{Alice Oh}.} \bibinfo{year}{2017}\natexlab{}.
\newblock \showarticletitle{Rotated Word Vector Representations and their
  Interpretability}. In \bibinfo{booktitle}{\emph{Proceedings of the 2017
  Conference on Empirical Methods in Natural Language Processing}}.
  \bibinfo{publisher}{Association for Computational Linguistics},
  \bibinfo{address}{Copenhagen, Denmark}, \bibinfo{pages}{401--411}.
\newblock
\urldef\tempurl%
\url{https://doi.org/10.18653/v1/D17-1041}
\showDOI{\tempurl}


\bibitem[\protect\citeauthoryear{Pennington, Socher, and Manning}{Pennington
  et~al\mbox{.}}{2014}]%
        {pennington:glove14}
\bibfield{author}{\bibinfo{person}{Jeffrey Pennington},
  \bibinfo{person}{Richard Socher}, {and} \bibinfo{person}{Christopher~D.
  Manning}.} \bibinfo{year}{2014}\natexlab{}.
\newblock \showarticletitle{{GloVe}: Global Vectors for Word Representation}.
  In \bibinfo{booktitle}{\emph{Proceedings of the 2014 Conference on Empirical
  Methods in Natural Language Processing}}. \bibinfo{pages}{1532--1543}.
\newblock
\urldef\tempurl%
\url{http://aclweb.org/anthology/D/D14/D14-1162.pdf}
\showURL{%
\tempurl}


\bibitem[\protect\citeauthoryear{Pitel and Fouquier}{Pitel and
  Fouquier}{2015}]%
        {DBLP:journals/corr/PitelF15}
\bibfield{author}{\bibinfo{person}{Guillaume Pitel} {and}
  \bibinfo{person}{Geoffroy Fouquier}.} \bibinfo{year}{2015}\natexlab{}.
\newblock \showarticletitle{Count-Min-Log sketch: Approximately counting with
  approximate counters}.
\newblock \bibinfo{journal}{\emph{CoRR}}  \bibinfo{volume}{abs/1502.04885}
  (\bibinfo{year}{2015}).
\newblock
\showeprint[arXiv]{1502.04885}
\urldef\tempurl%
\url{http://arxiv.org/abs/1502.04885}
\showURL{%
\tempurl}


\bibitem[\protect\citeauthoryear{Ribeiro, Singh, and Guestrin}{Ribeiro
  et~al\mbox{.}}{2016}]%
        {ribeiro:lime}
\bibfield{author}{\bibinfo{person}{Marco~Tulio Ribeiro},
  \bibinfo{person}{Sameer Singh}, {and} \bibinfo{person}{Carlos Guestrin}.}
  \bibinfo{year}{2016}\natexlab{}.
\newblock \showarticletitle{"Why should I trust you?" Explaining the
  predictions of any classifier}. In \bibinfo{booktitle}{\emph{Proceedings of
  the 22nd ACM SIGKDD international conference on knowledge discovery and data
  mining}}. \bibinfo{pages}{1135--1144}.
\newblock


\bibitem[\protect\citeauthoryear{Rinberg, Spiegelman, Bortnikov, Hillel,
  Keidar, and Serviansky}{Rinberg et~al\mbox{.}}{2019}]%
        {DBLP:journals/corr/abs-1902-10995}
\bibfield{author}{\bibinfo{person}{Arik Rinberg}, \bibinfo{person}{Alexander
  Spiegelman}, \bibinfo{person}{Edward Bortnikov}, \bibinfo{person}{Eshcar
  Hillel}, \bibinfo{person}{Idit Keidar}, {and} \bibinfo{person}{Hadar
  Serviansky}.} \bibinfo{year}{2019}\natexlab{}.
\newblock \showarticletitle{Fast Concurrent Data Sketches}.
\newblock \bibinfo{journal}{\emph{CoRR}}  \bibinfo{volume}{abs/1902.10995}
  (\bibinfo{year}{2019}).
\newblock
\showeprint[arXiv]{1902.10995}
\urldef\tempurl%
\url{http://arxiv.org/abs/1902.10995}
\showURL{%
\tempurl}


\bibitem[\protect\citeauthoryear{Rogers, Kovaleva, and Rumshisky}{Rogers
  et~al\mbox{.}}{2020}]%
        {bertology}
\bibfield{author}{\bibinfo{person}{Anna Rogers}, \bibinfo{person}{Olga
  Kovaleva}, {and} \bibinfo{person}{Anna Rumshisky}.}
  \bibinfo{year}{2020}\natexlab{}.
\newblock \showarticletitle{A Primer in BERTology: What we know about how
  {BERT} works}.
\newblock \bibinfo{journal}{\emph{CoRR}}  \bibinfo{volume}{abs/2002.12327}
  (\bibinfo{year}{2020}).
\newblock
\showeprint[arXiv]{2002.12327}
\urldef\tempurl%
\url{https://arxiv.org/abs/2002.12327}
\showURL{%
\tempurl}


\bibitem[\protect\citeauthoryear{Rudin}{Rudin}{2019}]%
        {rudin:intp}
\bibfield{author}{\bibinfo{person}{Cynthia Rudin}.}
  \bibinfo{year}{2019}\natexlab{}.
\newblock \showarticletitle{Stop explaining black box machine learning models
  for high stakes decisions and use interpretable models instead}.
\newblock \bibinfo{journal}{\emph{Nature Machine Intelligence}}
  (\bibinfo{date}{May} \bibinfo{year}{2019}), \bibinfo{pages}{206--215}.
\newblock


\bibitem[\protect\citeauthoryear{{State of California}}{{State of
  California}}{2021}]%
        {ca-toxicity-data}
\bibfield{author}{\bibinfo{person}{{State of California}}.}
  \bibinfo{year}{2021}\natexlab{}.
\newblock \bibinfo{title}{Surface Water - Toxicity Results}.
\newblock
\newblock
\urldef\tempurl%
\url{https://data.cnra.ca.gov/dataset/surface-water-toxicity-results}
\showURL{%
\tempurl}


\bibitem[\protect\citeauthoryear{Subramanian, Pruthi, Jhamtani,
  Berg{-}Kirkpatrick, and Hovy}{Subramanian et~al\mbox{.}}{2017}]%
        {SPINE}
\bibfield{author}{\bibinfo{person}{Anant Subramanian}, \bibinfo{person}{Danish
  Pruthi}, \bibinfo{person}{Harsh Jhamtani}, \bibinfo{person}{Taylor
  Berg{-}Kirkpatrick}, {and} \bibinfo{person}{Eduard~H. Hovy}.}
  \bibinfo{year}{2017}\natexlab{}.
\newblock \showarticletitle{{SPINE:} SParse Interpretable Neural Embeddings}.
\newblock \bibinfo{journal}{\emph{CoRR}}  \bibinfo{volume}{abs/1711.08792}
  (\bibinfo{year}{2017}).
\newblock
\showeprint[arXiv]{1711.08792}
\urldef\tempurl%
\url{http://arxiv.org/abs/1711.08792}
\showURL{%
\tempurl}


\bibitem[\protect\citeauthoryear{Thomas, Bordawekar, Aggarwal, and Yu}{Thomas
  et~al\mbox{.}}{2009}]%
        {thomas:cell}
\bibfield{author}{\bibinfo{person}{Dina Thomas}, \bibinfo{person}{Rajesh
  Bordawekar}, \bibinfo{person}{Charu~C. Aggarwal}, {and}
  \bibinfo{person}{Philip~S. Yu}.} \bibinfo{year}{2009}\natexlab{}.
\newblock \showarticletitle{On Efficient Query Processing of Stream Counts on
  the Cell Processor}. In \bibinfo{booktitle}{\emph{2009 IEEE 25th
  International Conference on Data Engineering}}. \bibinfo{pages}{748--759}.
\newblock
\urldef\tempurl%
\url{https://doi.org/10.1109/ICDE.2009.35}
\showDOI{\tempurl}


\bibitem[\protect\citeauthoryear{{UCI Data Team}}{{UCI Data Team}}{2019}]%
        {mashroom-data}
\bibfield{author}{\bibinfo{person}{{UCI Data Team}}.}
  \bibinfo{year}{2019}\natexlab{}.
\newblock \bibinfo{title}{UCI Mashroom Data}.
\newblock
\newblock
\urldef\tempurl%
\url{https://www.kaggle.com/code/aavigan/uci-mushroom-data}
\showURL{%
\tempurl}


\bibitem[\protect\citeauthoryear{van~der Maaten and Hinton}{van~der Maaten and
  Hinton}{2008}]%
        {JMLR:v9:vandermaaten08a}
\bibfield{author}{\bibinfo{person}{Laurens van~der Maaten} {and}
  \bibinfo{person}{Geoffrey Hinton}.} \bibinfo{year}{2008}\natexlab{}.
\newblock \showarticletitle{Visualizing Data using t-SNE}.
\newblock \bibinfo{journal}{\emph{Journal of Machine Learning Research}}
  \bibinfo{volume}{9}, \bibinfo{number}{86} (\bibinfo{year}{2008}),
  \bibinfo{pages}{2579--2605}.
\newblock
\urldef\tempurl%
\url{http://jmlr.org/papers/v9/vandermaaten08a.html}
\showURL{%
\tempurl}


\bibitem[\protect\citeauthoryear{Vaswani, Shazeer, Parmar, Uszkoreit, Jones,
  Gomez, Kaiser, and Polosukhin}{Vaswani et~al\mbox{.}}{2017}]%
        {vaswani2017attention}
\bibfield{author}{\bibinfo{person}{Ashish Vaswani}, \bibinfo{person}{Noam
  Shazeer}, \bibinfo{person}{Niki Parmar}, \bibinfo{person}{Jakob Uszkoreit},
  \bibinfo{person}{Llion Jones}, \bibinfo{person}{Aidan~N. Gomez},
  \bibinfo{person}{Lukasz Kaiser}, {and} \bibinfo{person}{Illia Polosukhin}.}
  \bibinfo{year}{2017}\natexlab{}.
\newblock \bibinfo{title}{Attention Is All You Need}.
\newblock
\newblock
\showeprint[arxiv]{1706.03762}~[cs.CL]


\bibitem[\protect\citeauthoryear{Wiegreffe and Pinter}{Wiegreffe and
  Pinter}{2019}]%
        {DBLP:journals/corr/abs-1908-04626}
\bibfield{author}{\bibinfo{person}{Sarah Wiegreffe} {and}
  \bibinfo{person}{Yuval Pinter}.} \bibinfo{year}{2019}\natexlab{}.
\newblock \showarticletitle{Attention is not not Explanation}.
\newblock \bibinfo{journal}{\emph{CoRR}}  \bibinfo{volume}{abs/1908.04626}
  (\bibinfo{year}{2019}).
\newblock
\showeprint[arXiv]{1908.04626}
\urldef\tempurl%
\url{http://arxiv.org/abs/1908.04626}
\showURL{%
\tempurl}


\bibitem[\protect\citeauthoryear{Zhang, Du, Sun, and Li}{Zhang
  et~al\mbox{.}}{2019}]%
        {zhang:w2v}
\bibfield{author}{\bibinfo{person}{Haitong Zhang}, \bibinfo{person}{Yongping
  Du}, \bibinfo{person}{Jiaxin Sun}, {and} \bibinfo{person}{Qingxiao Li}.}
  \bibinfo{year}{2019}\natexlab{}.
\newblock \showarticletitle{Improving Interpretability of Word Embeddings by
  Generating Definition and Usage}.
\newblock \bibinfo{journal}{\emph{CoRR}}  \bibinfo{volume}{abs/1912.05898}
  (\bibinfo{year}{2019}).
\newblock
\showeprint[arXiv]{1912.05898}
\urldef\tempurl%
\url{http://arxiv.org/abs/1912.05898}
\showURL{%
\tempurl}


\bibitem[\protect\citeauthoryear{Zhang, Ti{\~{n}}o, Leonardis, and Tang}{Zhang
  et~al\mbox{.}}{2020}]%
        {zhang:survey}
\bibfield{author}{\bibinfo{person}{Yu Zhang}, \bibinfo{person}{Peter
  Ti{\~{n}}o}, \bibinfo{person}{Ales Leonardis}, {and} \bibinfo{person}{Ke
  Tang}.} \bibinfo{year}{2020}\natexlab{}.
\newblock \showarticletitle{A Survey on Neural Network Interpretability}.
\newblock \bibinfo{journal}{\emph{CoRR}}  \bibinfo{volume}{abs/2012.14261}
  (\bibinfo{year}{2020}).
\newblock
\showeprint[arXiv]{2012.14261}
\urldef\tempurl%
\url{https://arxiv.org/abs/2012.14261}
\showURL{%
\tempurl}


\end{thebibliography}

\end{document}